\documentclass{article} %

\PassOptionsToPackage{dvipsnames}{xcolor}

\usepackage{iclr2027_conference,times}

\usepackage{amsmath,amsfonts,bm}

\def\eqref#1{equation~\ref{#1}}

\def\1{\bm{1}}

\DeclareMathAlphabet{\mathsfit}{\encodingdefault}{\sfdefault}{m}{sl}
\SetMathAlphabet{\mathsfit}{bold}{\encodingdefault}{\sfdefault}{bx}{n}

\newcommand{\R}{\mathbb{R}}

\usepackage[utf8]{inputenc}      %
\usepackage[T1]{fontenc}         %
\usepackage{url}
\usepackage{graphicx}
\usepackage{booktabs}            %
\usepackage{multirow}
\usepackage{microtype}           %
\usepackage{xcolor}              %
\usepackage{colortbl}
\usepackage{adjustbox}
\usepackage{placeins}
\usepackage{enumitem}
\usepackage{xspace}
\usepackage{hyperref}
\usepackage{caption}             %
\usepackage[capitalize]{cleveref}

\newcommand{\ours}{WarpHammer\xspace}

\newcommand{\B}[1]{\textbf{#1}}
\renewcommand{\R}{\mathbb{R}}   %
\newcommand{\C}{\mathcal{C}}    %
\newcommand{\dg}[1]{$^{\textcolor{ForestGreen}{\scriptsize#1}}$}

\title{WarpHammer: Densifying Scene Warps with 3D \\ Object Priors for Extreme View Synthesis}

\author{
\makebox[\textwidth][c]{%
\textbf{Michael Green}$^{1,4}$,
\textbf{Gavriel Habib}$^{1}$,
\textbf{Dvir Samuel}$^{2}$,
\textbf{Tal Berkovitz Shalev}$^{1}$%
}\\[-0.1em]
\makebox[\textwidth][c]{%
\textbf{Issar Tzachor}$^{1}$,
\textbf{Rami Ben-Ari}$^{1,*}$,
\textbf{Or Litany}$^{3,2,*}$%
}\\[0.4em]
\makebox[\textwidth][c]{%
$^{1}$OriginAI, Israel
\qquad
$^{2}$NVIDIA
\qquad
$^{3}$Technion
\qquad
$^{4}$General Motors, Herzliya, Israel%
}\\[0.3em]
\makebox[\textwidth][c]{%
{\small $^{*}$Authors jointly supervised this work.}%
}
}

\iclrfinalcopy

\begin{document}

\maketitle
\lhead{Preprint}

\begin{center}
\textbf{Project page:}
\href{https://pihash2k.github.io/extreme_nvs/}
{\texttt{pihash2k.github.io/extreme\_nvs}}
\end{center}

\begin{abstract}
Projection-conditioned novel view synthesis (NVS) warps an explicit 3D reconstruction of the input view into the target camera and conditions a generator on the warped rendering. This works well for small viewpoint changes but degrades sharply under large orbital motion: the warp becomes sparse around the orbited object, where hidden surfaces dominate the new view and mirror-like artifacts emerge, causing the generator to lose both pixel content and the implicit camera cue carried by the warp. We introduce \ours{}, a training-free framework that resolves this failure mode by augmenting the warped scene with an explicit 3D reconstruction of the object obtained from a native 3D generative prior (e.g., SAM3D). The reconstructed object adds missing foreground surfaces and occludes background points that should no longer be visible, restoring both appearance and camera cues without fine-tuning the base model. The same explicit object representation further unlocks a capability current NVS pipelines do not support: incorporating auxiliary views of the object from sources outside the target scene, for example, a casual snapshot of a car paired with a manufacturer studio shot of the same model. We process the reference and auxiliary images jointly with a pretrained multi-view geometry foundation model, which predicts a unified point cloud that we fuse into the 3D object reconstruction. This yields substantially more faithful geometry than single-image reconstruction, without requiring user-provided camera poses for the auxiliary views. On five benchmarks, \ours{} produces stable novel views at viewpoint deviations where strong baselines collapse, and is the first scene-level NVS method that can naturally fuse auxiliary, unposed object views from an external source, recovering their relative pose rather than requiring it as input.

\end{abstract}

\section{Introduction}
Novel view synthesis (NVS) from a single image or a sparse set of views is fundamentally underconstrained: the input does not contain sufficient evidence to recover unobserved regions of the scene. Recent work has therefore shifted from purely per-scene reconstruction to learned visual priors. These include pretrained image generators~\citep{rombach2022stable,podell2023sdxl}, multi-view generative models~\citep{liu2023zero123,liu2023syncdreamer,long2024wonder3d,seva}, and video generators~\citep{blattmann2023stable,chen2023videocrafter,liang2026orbitnvs}. Trained on large-scale visual data, these models learn statistical regularities that enable plausible completion of missing scene content.

There are two dominant paradigms in NVS. Camera-conditioned methods encode the target camera directly, to infer scene structure implicitly~\citep{lvsm,cat3d,cat4d,seva,rayzer}. Projection-conditioned methods instead construct an explicit 3D proxy from the input, render it from the target viewpoint, and condition generation on the resulting warp~\citep{gen3c,viewcrafter,ex4d,flexworld,see3d,yu2025trajectorycrafter,streetcrafter,mvgenmaster,gsenhancer,splatdiff}. This explicit geometric conditioning enables strong camera control and state-of-the-art NVS performance~\citep{gen3c,ex4d}. We build on this paradigm.  

However, both paradigms remain brittle under large viewpoint extrapolation. Camera-conditioned models can fail due to out-of-distribution motion~\citep{raystoprojections}, while projection-conditioned methods introduce artifacts since the warped point cloud develops holes and sparse regions as the view angle increases~\citep{closeupshot,fixAnything2026}.

We focus primarily on object-anchored scenes,  where a foreground object constitutes the main subject of the image and therefore more clearly determines the perceptual fidelity of the synthesized view. Such scenes are ubiquitous in product visualization, object-centric capture, and content creation. They are also where large viewpoint changes matter most, and such changes are particularly challenging here because substantial portions of the foreground object are unobserved in the input.

\begin{figure}[!htbp]
    \centering
\includegraphics[width=0.85\linewidth, trim={0 4.5cm 0 4.5cm}, clip]{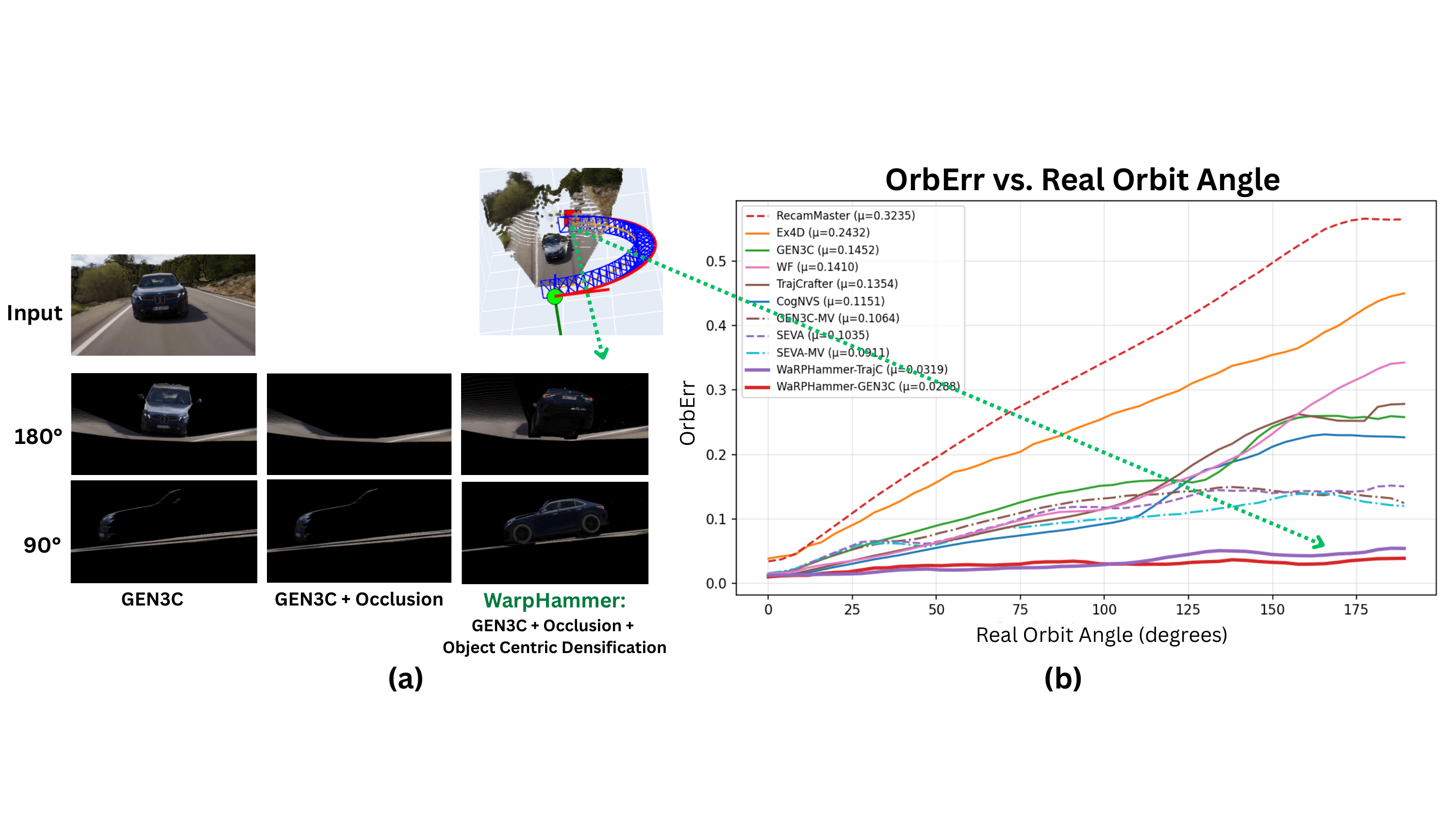}
\captionsetup{font=small}
\caption{Object-centric densification improves extreme-view synthesis.
\textbf{(a)~Qualitative:}~GEN3C shows sparsity and artifacts; \ours{} remains geometry-consistent.
\textbf{(b)~Quantitative:}~OrbErr vs.\ angle: baselines degrade beyond ${\sim}110^{\circ}$, while \ours{} maintains low error.}
    \label{fig:rot_error_yaw}
\end{figure}

Importantly, the projected warp also acts as a geometric scaffold that communicates the prescribed camera motion to the generator~\citep{fixAnything2026}. This provides an advantage over camera-conditioned image/video generative models, which often produce a plausible or likely camera motion rather than faithfully following the user-specified trajectory~\citep{fixAnything2026}. Consequently, when the warp becomes severely sparse at extreme viewpoints, the generator loses both appearance evidence and reliable geometric guidance, and pose and reconstruction errors increase rapidly under large orbital motion (Fig.~\ref{fig:rot_error_yaw}). Yet this failure is not directly measured by standard camera-control metrics: rotation error, for example, can remain low even when the object is rendered in an incorrect orientation. To quantify it, we introduce OrbErr, an orbital-fidelity metric for object-centric NVS (Appendix~\ref{sec:supp_orberr}).

We introduce \ours{}, a training-free framework that mitigates this failure mode particularly on object-anchored scenes, by densifying the warped conditioning signal with an explicit object-centric 3D prior. Our key observation is that the scene warp is most unreliable exactly where object geometry matters most: on the newly visible surfaces of the central object. We augment the standard scene cache with a reconstructed 3D object from a native 3D generative model. The reconstructed mesh plays a dual role: it adds geometry for newly visible object surfaces, and it occludes background points that should no longer be visible once the camera has orbited past the object, thus eliminating the background leakage that scene-only warps suffer from at large angles.

A key aspect of our design is the separation between geometry and appearance. Native 3D generators (e.g., Hunyuan3D, TRELLIS) provide reliable shape but their texture and illumination may deviate from the reference scene, and they often lack photorealism~\citep{sobol2026realiz3d}. We therefore use the reconstruction as geometric guidance only, leaving the diffusion model's scene-identity conditioning responsible for appearance.

Beyond improving single-image extreme-view synthesis, the explicit object representation enables a second capability that standard warp-based pipelines do not naturally support: incorporating an auxiliary object view from outside the target scene. For example, a dashcam image may contain a car from one viewpoint, while a catalog or manufacturer image may show the same or a related car from another viewpoint. Such external views can contain object-specific information that a generative prior cannot infer from the reference alone, such as the shape of unobserved surfaces, a logo, or a license plate.  These auxiliary images are typically unposed and may be visually mismatched with the reference scene. We jointly process the reference and auxiliary object views with a pretrained multi-view geometry foundation model, which aligns the auxiliary geometry to the reference object without camera calibration from the user. The generated video remains anchored to the appearance and context of the reference scene; when the auxiliary appearance is mismatched, only its geometry is used (\S\ref{sec:method_aux}). 

Our contributions are: 
\begin{itemize}[leftmargin=*]
    \item We substantially strengthen extreme-view synthesis in existing projection-conditioned scene NVS methods without additional training, by densifying sparse target-view warps with an explicit object-centric 3D reconstruction.        
    \item We define and address an auxiliary-view NVS setting, where an unposed object image from an external source is fused into the reference scene as structural evidence rather than as a calibrated scene view. 
    \item We propose an appearance-aware fusion strategy that separates object geometry from appearance, enabling auxiliary views to improve structure without forcing inconsistent color, texture, or illumination into the output. We also introduce \textbf{SynView-X}, a benchmark for object-centric NVS under large viewpoint changes. 
    \item We introduce OrbErr, an orbital-fidelity metric for object-centric NVS, together with a benchmark for unposed object-view fusion, and evaluate across single-view and auxiliary-view regimes on five datasets.
\end{itemize}

\section{Related Work}
\paragraph{Sparse-view and multi-view NVS.}
Classical NVS methods such as NeRFs~\citep{mildenhall2020nerf,mipnerf360} and 3D Gaussian Splatting~\citep{kerbl2023gaussian} achieve high-quality rendering from dense posed views, while generalizable methods reduce the number of inputs using learned priors~\citep{yu2021pixelnerf,wang2021ibrnet,zhou2023sparsefusion}. However, these methods still assume views from the same calibrated scene. \ours{} instead allows an auxiliary object image to be unposed, out-of-scene, and appearance-mismatched, treating it as structural evidence rather than as another calibrated scene view.

\paragraph{Object-centric generative view synthesis.}
A parallel line of work uses generative priors for single-image object view synthesis. Zero-1-to-3~\citep{liu2023zero123} learns camera-conditioned novel-view generation from a single object image. SyncDreamer~\citep{liu2023syncdreamer} and Wonder3D~\citep{long2024wonder3d} improve multi-view consistency by jointly generating synchronized views, normals, or other geometry-aware intermediates. SV3D~\citep{voleti2024sv3d} extends this idea with video diffusion and explicit orbital camera control, while SV4D~\citep{xie2025sv4d} produces multi-frame, multi-view consistent videos for dynamic objects. Zero-to-Hero~\citep{sobol2024zero} further improves single-image NVS at challenging viewpoints through training-free attention-map manipulation at inference.
These methods demonstrate that generative models encode strong object-level 3D priors, but they typically synthesize isolated objects on clean backgrounds and stop short
of integrating objects back into a real reference scene with its original lighting and surroundings. \ours{} instead uses object-centric 3D generation as structural guidance inside a scene-level NVS pipeline, while appearance, lighting, and background context remain anchored to the reference image.

\paragraph{Camera-controlled and projection-conditioned generative NVS.}
Recent video diffusion models enable camera-controllable generation from images or monocular videos by injecting camera motion through learned pose embeddings, Pl\"ucker rays, relative position encodings, or dedicated control branches~\citep{wang2024motionctrl,he2025cameractrl,bahmani2025vd3d,bahmani2025ac3d,li2025realcami2v,luo2025camclonemaster,he2025cameractrl2}. These approaches provide flexible camera control, but the camera signal is mostly implicit, and object geometry may drift under large viewpoint changes.

Projection-conditioned methods make the camera conditioning more explicit. They lift the source image or video into a 3D proxy, render it from the target camera, and use the resulting warp to guide a diffusion model. ViewCrafter~\citep{viewcrafter} and TrajectoryCrafter~\citep{yu2025trajectorycrafter} follow this strategy for single-image and monocular-video inputs, respectively, while GEN3C~\citep{gen3c} maintains a persistent 3D cache for world-consistent generation. SEVA~\citep{seva} accepts posed input views and target cameras, and related methods explore learned camera encoders or inference-time motion guidance~\citep{zhang2025spatialcrafter,song2025worldforge}. 
However, these methods ultimately condition on geometry visible in the input view. As the target camera moves to large angles, hidden object surfaces must be hallucinated by the diffusion prior. \ours{} addresses this limitation by augmenting the scene warp with an explicit object-centric 3D reconstruction, optionally strengthened by an auxiliary object view.

\paragraph{Geometry-completed proxies for large-viewpoint synthesis.}
Several recent methods also improve large-viewpoint synthesis by constructing more complete geometric proxies. CogNVS~\citep{chen2025cognvs} reconstructs co-visible regions from monocular video, inpaints hidden regions, and applies test-time finetuning for temporal coherence. EX-4D~\citep{ex4d} uses a depth-watertight mesh as guidance for extreme-viewpoint 4D synthesis. Concurrent work FreeOrbit4D~\citep{freeorbit4d} builds on VACE~\citep{jiang2025vace} depth conditioning to complete foreground and background geometry for arbitrary camera redirection in monocular videos. 
These methods share a similar motivation, but primarily derive the proxy from monocular video or depth and target dynamic or 4D content.
\ours{} targets single-image static-scene NVS, using a native 3D generative prior for object geometry and optionally refining it with an unposed external object image.

\section{Method}

\begin{figure*}[t]
    \centering
    \includegraphics[width=\textwidth]{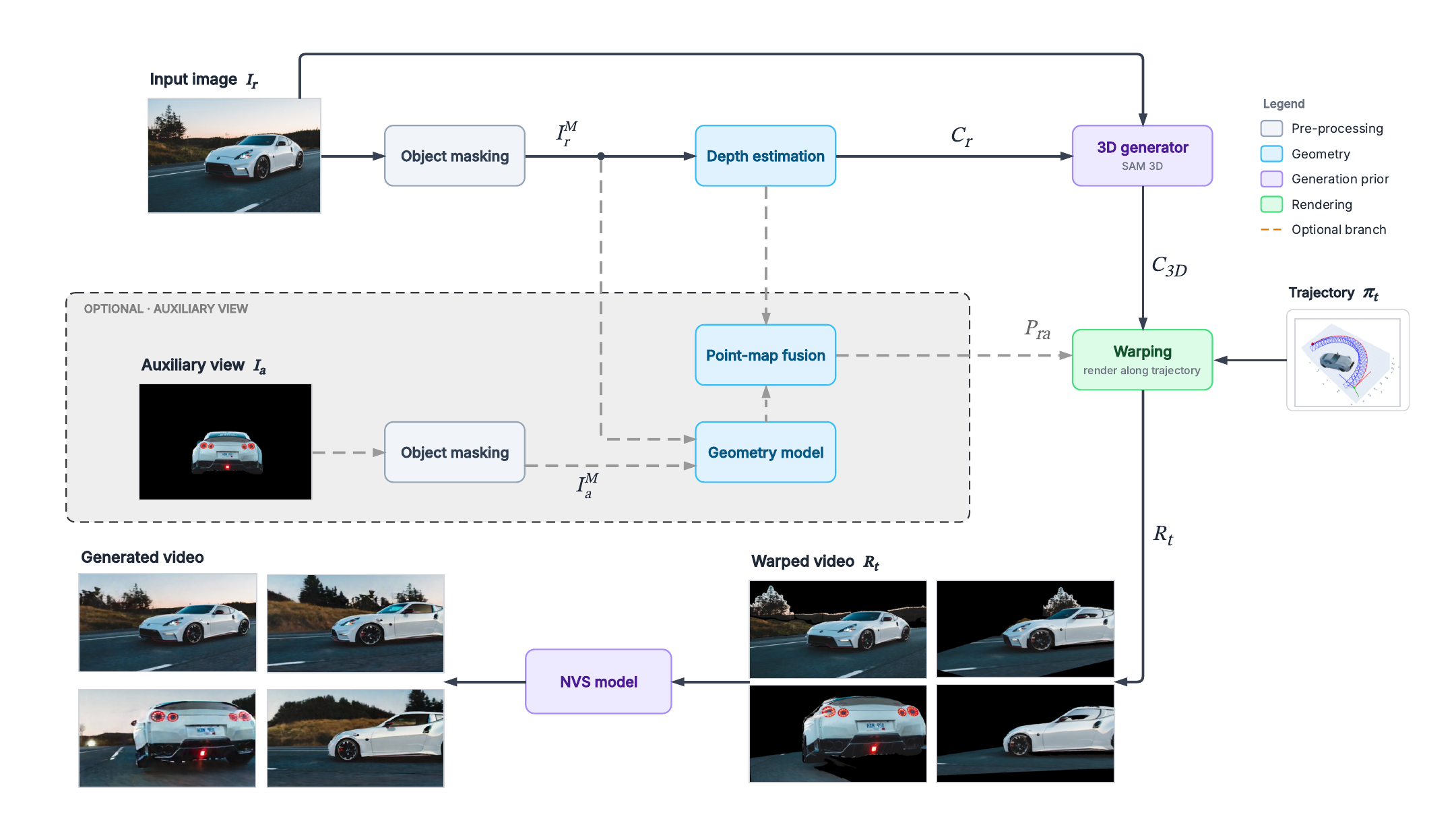}
    \captionsetup{font=small}
    \caption{
\textbf{Overview of \ours{}.}
From a reference image $I_r$ and target trajectory $\{\pi_t\}_{t=1}^T$ (and an optional auxiliary object image $I_a$), \ours{} builds a densified point-cloud cache by augmenting the scene warp with an explicit 3D object reconstruction, optionally fused with $I_a$ via a multi-view geometry foundation model, and renders it to condition a video diffusion prior.
    }

    \label{fig:method}
\end{figure*}

\Cref{fig:method} gives an overview of the pipeline. We first describe the single-image pipeline (\S\ref{sec:method_prelim}--\S\ref{sec:method_densify}), then show how the same machinery extends to an auxiliary, unposed object view from a different scene (\S\ref{sec:method_aux}).

\paragraph{Problem setup.} We assume the reference scene image $I_r \in \mathbb{R}^{H \times W \times 3}$ contains a dominant foreground object, the \emph{central object}, around which the target camera trajectory orbits. We further assume that the central object is rigid, that an off-the-shelf segmenter isolates it, and that enough of it is visible in $I_r$ for the 3D prior to reconstruct it.  Given $I_r$ and a 6-DoF camera trajectory $\{\pi_t\}_{t=1}^T$ specified \emph{relative} to the input view, our goal is to synthesize a novel-view video $V_g$ of $T$ RGB frames at resolution $H \times W$ that follows the trajectory while preserving scene appearance and object identity. \S\ref{sec:method_aux} extends this setup with an additional, uncalibrated auxiliary image $I_a$ of the same object, possibly captured in a different scene and under different illumination.

\subsection{Preliminaries: Projection-Conditioned NVS via Scene Caches}
\label{sec:method_prelim}

We build \ours{} on top of two representative backbones: GEN3C~\citep{gen3c} and TrajectoryCrafter~\citep{yu2025trajectorycrafter}, both of which lift the reference image into a point cloud via monocular depth estimation, render it along the target trajectory, and condition a video diffusion prior on the resulting warped buffers. We describe the lifted geometry abstractly as a {\it cache}
\begin{equation}
    \C = \{P, A\}, \qquad P \subset \R^3, \;\; A \in \R^{|P|\times 3},
    \label{eq:cache}
\end{equation}

where $P$ are the lifted scene points and $A$ their RGB attributes; trajectory poses $\{\pi_t\}$ are expressed relative to the reference camera. Rendering $\C$ along $\{\pi_t\}_{t=1}^T$ produces warped renderings $\{R_t\}_{t=1}^T$ and per-pixel confidence masks $\{M_t\}_{t=1}^T$, $M_t\in[0,1]^{H\times W}$, which are binary validity masks in the base pipelines, that anchor synthesized frames to the input scene while leaving the diffusion prior to fill newly visible regions. The video diffusion prior $\mathcal{G}$ maps these buffers, together with $I_r$, to the output video, $V_g=\mathcal{G}(I_r,\{R_t,M_t\}_{t=1}^T)$.

\subsection{Densifying the Object with a 3D Object Prior}
\label{sec:method_densify}

The object is where sparsity hurts most, and where it matters most: background regions typically remain supported as the camera translates, but the object's hidden surfaces become dominant the moment the camera orbits, and these are precisely the surfaces that define its identity. We address this directly by densifying the foreground cache with an explicit 3D reconstruction from a point-map-conditioned native object generator~\citep{sam3d}
which recovers a complete object representation from a single image. 

\paragraph{Object reconstruction and placement.}
We isolate the central foreground object in $I_r$ using an off-the-shelf segmenter, producing masked image $I_r^M$. We then pass $I_r^M$ together with the masked object points $P_r$ of the reference cache $\C_r$ (\Cref{eq:cache}) to a point-cloud-conditioned 3D generator $f_{3D}$, instantiated with SAM3D~\citep{sam3d}, an image-to-3D object generator that accepts a point map as conditioning and returns a textured mesh aligned to it; we denote the resulting object geometry by $P_{3D}=f_{3D}(I_r^M,P_r)$. SAM3D predicts the object's pose and scale in the frame of its conditioning point map, so the mesh is returned directly in the reference camera frame at the scale of $P_r$, with no separate registration step.
Other image-to-3D generators can also serve as $f_{3D}$ (Appendix~\ref{sec:supp_generator}). 

\paragraph{Cache fusion and visibility.}
We form the fused cache by inserting the densified object into the reference cache,
\begin{equation}
    \C_f \;=\; \C_r \;\cup\; \C_{3D},
    \qquad \C_{3D} = \{P_{3D}\},
    \label{eq:cache_fusion}
\end{equation}
where $\C_{3D}$ carries only the reconstructed geometry and no color attributes, and resolve visibility during rendering. Because SAM3D natively returns a mesh, we render the densified object portion of $\C_f$ from that mesh instead of splatting the sampled $P_{3D}$ as points; its conditioning values are defined below. We splat $\C_r$ and rasterize the mesh into a shared depth buffer, and discard a reference point when its depth exceeds the mesh depth along the same ray by more than a small tolerance. This removes stale reference points that should be hidden once the camera has orbited past the object.

Because of this occlusion, reconstruction errors do more than add incorrect geometry: a misplaced surface can also remove correct reference points behind it. The warped video $\{R_t\}$ and mask $\{M_t\}$ are produced jointly from $\C_f$ and handed to the diffusion prior as in the base pipeline, with the foreground region now densely supported across the full trajectory.

\paragraph{Geometry-focused conditioning.}
While SAM3D yields reliable shape estimates, its texture and color can deviate from the reference scene's illumination and appearance and generally fall short of the reference image's photorealism. We therefore use the reconstructed mesh without its color. Let $\Omega_t$ be the pixels of target frame $t$ that the mesh covers but no warped reference point does, and let $\hat{D}_t$ be the depth rendered from the mesh. Within $\Omega_t$,
\begin{equation}
    R_t(u) = g\big(\hat{D}_t(u)\big),
    \label{eq:geo_cond}
\end{equation}
where $g$ maps the rendered depth to a normalized gray value replicated across the three color channels (Appendix~\ref{sec:supp_aux_neutral}). The depth code is thus written into the rendered RGB conditioning frames $\{R_t\}$, not into the cache attributes $A$, so that the densified region carries geometric structure but no donor-specific color identity. The diffusion prior, conditioned on $I_r$, then fills in the appearance of these regions, keeping it coherent with the reference scene.
When no foreground object is available for augmentation,
$\C_{3D}=\emptyset$ and $\C_f=\C_r$, so our formulation reduces
to the underlying projection-conditioned pipeline.

\subsection{Fusing an Auxiliary Object View}
\label{sec:method_aux}
Beyond single-image extreme-view synthesis, the object representation from \S\ref{sec:method_densify} also lets \ours{} incorporate an uncalibrated auxiliary image $I_a$ of the same object, even when it is captured outside the reference scene, e.g., from a catalog, studio image, or web retrieval.

Since $I_a$ has no known pose relative to $I_r$ and may differ in background, lighting, color, or texture, \ours{} uses it as additional \emph{geometric evidence} rather than as a calibrated second view; whether its RGB is retained depends on the appearance match, as described below.

\paragraph{Joint geometry from an unposed auxiliary view.}
We mask the auxiliary image to the object, $I_a^M$, and pass it, together with the masked reference $I_r^M$, to a feed-forward multi-view geometry model $f_g$, instantiated with VGGT~\citep{vggt}:
\begin{equation}
    [P_r^g,\,P_a^g] \;=\; f_g(I_r^M,\, I_a^M).
\end{equation}
The model jointly predicts dense object pointmaps $P_r^g, P_a^g$ in a shared coordinate frame, so the auxiliary pointmap arrives already registered to the reference object, but in VGGT's own frame and scale. Masking the inputs to $f_g$ is important in practice: it focuses the model on object surfaces and stabilizes the predicted relative pose for wide baselines, including hard front-to-back configurations.

\paragraph{Cross-view point-map fusion.}
We keep confidence-filtered points and fuse the auxiliary geometry with the reference cache by mapping it into the reference cache's frame. The anchor is the reference object, which is observed twice: as $P_r$, lifted by the model's depth, and as $P_r^g$, predicted by VGGT. Both are indexed by the same reference pixels, so they provide direct 3D correspondences, from which we estimate a similarity transform (scale $s$, rotation $\mathbf{R}$, translation $\mathbf{t}$) in closed form (Appendix~\ref{sec:supp_alignment}):
\begin{equation}
    \hat{P}_a \;=\; s\,\mathbf{R}\,P_a^g+\mathbf{t},
    \qquad
    P_{ra} \;=\; P_r \,\cup\, \hat{P}_a.
    \label{eq:aux_align}
\end{equation}
We call $P_{ra}$ the \emph{fused object point map}: it combines the surfaces visible in the reference view ($P_r$) with those contributed by the auxiliary view ($\hat{P}_a$), in a single frame and scale.

\paragraph{Aligned 3D generation with auxiliary support.}
The fused object point map replaces $P_r$ as the conditioning of $f_{3D}$ in the densification step,
\begin{equation}
    P_{3D} \;=\; f_{3D}(I_r^M,\, P_{ra}),
\end{equation}
where $\hat{P}_a$ supplies real evidence for surfaces that are weakly observed or absent from the reference view. The fused cache becomes $\C_f = \C_r \cup \C_{3D} \cup \C_a$, with the auxiliary cache $\C_a = \{\hat{P}_a, A_a\}$ also expressed in the reference frame; rendering and conditioning then proceed exactly as in \S\ref{sec:method_densify}.

\paragraph{Appearance-neutral auxiliary conditioning.}
The auxiliary view reuses the depth code $g$ of \Cref{eq:geo_cond}, with $\Omega_t$ given by the footprint of its aligned geometry projected into each target view. Unlike the mesh texture, which is never used, auxiliary RGB is replaced by this depth code only when a configuration flag enables neutralization, i.e., when the donor color is not to be trusted; in that case, the mask inside the footprint is set to a constant $\alpha<1$ instead of full validity, so that the prior uses the depth code as a structural hint rather than reproducing it as gray appearance. Otherwise, auxiliary RGB and its validity mask are retained. Pixels outside the footprint fall back to the reference scene cache with zero auxiliary confidence. Appendix~\ref{sec:supp_aux_neutral} and~\ref{sec:supp_neutral_ablation} define these buffers and measure the suppression of donor color."

\section{Experiments}
\label{sec:evaluation}
We compare \ours{} against TrajectoryCrafter~\citep{yu2025trajectorycrafter}, 
GEN3C~\citep{gen3c}, ReCamMaster~\citep{bai2025recammaster}, EX-4D~\citep{ex4d}, 
CogNVS~\citep{chen2025cognvs}, SEVA~\citep{seva}, and WorldForge~\citep{song2025worldforge}, spanning 
projection-conditioned and camera-conditioned paradigms. 
FreeOrbit4D~\citep{freeorbit4d} shares a similar idea of completing the
geometric proxy, but accepts only video as input; for completeness, we
therefore compare with it qualitatively only (Appendix~\ref{sec:supp_freeorbit4d}).
For comparison with posed multi-view conditioning, we use the multi-view variants of GEN3C and SEVA, which accept a second input view \emph{with known pose from the same scene}, denoted 
GEN3C-MV and SEVA-MV, and report them as stronger-supervision references with GT 
camera poses.

\paragraph{Benchmarks.}
We evaluate on five benchmarks spanning synthetic, scanned, and real-world 
captured data. We introduce SynView-X, a benchmark for object-centric 
NVS under large orbital viewpoint changes, constructed from CAD assets drawn 
from ShapeNet~\citep{5shapenet} and Objaverse~\citep{objaverse}, augmented with 
scanned objects from GSO~\citep{gso}. Existing benchmarks do not specifically 
target extreme orbital deviations around a dominant foreground object. 
SynView-X fills this gap by providing ground-truth multi-view sequences with 
controlled large-angle coverage. For scanned and real-world evaluation we use OmniObject3D~\citep{wu2023omniobject3d}, 
Co3D~\citep{reizenstein2021common}, Objectron~\citep{ahmadyan2021objectron}, and 
Mip-NeRF~360~\citep{mipnerf360}. For each dataset, evaluation pairs are 
constructed by selecting images with large orbital viewpoint differences; one 
serves as the reference input and the other as the target view. Dataset details 
and construction protocol are provided in Appendix~\ref{sec:supp_datasets}.

\paragraph{Metrics.}
Our evaluation measures three complementary aspects of performance: image fidelity, geometric accuracy, and perceptual video quality. For image fidelity, we evaluate PSNR, SSIM, and LPIPS on the masked object region at the target viewpoint, and CLIP-F~\citep{bai2025recammaster} for 
semantic alignment with the target view, focusing the evaluation on the 
synthesized object rather than potentially unrelated background, where applicable (details 
in Appendix~\ref{sec:supp_metrics} and~\ref{sec:supp_quant}).

For geometric accuracy, we propose OrbErr, a metric tailored to 
object-centric orbital trajectories. A faithful orbit must satisfy two 
conditions simultaneously: the camera must follow the intended orbital 
viewing direction, and the rendered object must place its visual features at 
spatial locations consistent with that viewpoint. OrbErr therefore combines 
two complementary terms:
\begin{equation}
E_{\mathrm{orb}}(t) =
\tfrac{1}{2} E_{\mathrm{euler}}(t)
\;+\;
\tfrac{1}{2} E_{\mathrm{disp}}(t),
\end{equation}
where $E_{\mathrm{euler}}$ measures normalized wrapped angular deviation in 
azimuth and elevation from monocular pose estimates~\citep{lin2025depth}, and 
$E_{\mathrm{disp}}$ measures object-region feature displacement using mutual 
nearest-neighbor matches between DINOv2~\citep{oquab2023dinov2} patch features 
of the rendered and ground-truth frames. The displacement of these 
mutual matches is combined with a coverage penalty, preventing the metric from 
rewarding methods that match only a small subset of easy patches. Reporting 
the orbital-angle and feature-displacement terms jointly penalizes partial 
failures that rotation error or PSNR alone misses, including incorrect object 
orientation, mirror-like duplications, and incomplete object reconstruction. 

\subsection{Results}
\paragraph{Quantitative.}
Table~\ref{tab:nvs_mean_sv_main} reports single-view results 
averaged across all five benchmarks. Both \ours{}-GEN3C and 
\ours{}-TC improve PSNR, SSIM, and LPIPS over their backbones,
and achieve the lowest SynView-X OrbErr. For completeness, \Cref{tab:nvs_mean_sv_vbench} in the Appendix reports VBench~\citep{huang2024vbench} video-quality metrics.

Table~\ref{tab:nvs_mean_mv_main} reports results in the auxiliary-view
setting. Without any ground-truth camera poses or a full-scene second
image, \ours{}-GEN3C outperforms all baselines without GT poses and
approaches the image similarity of the GT-pose variants, which use
in-scene views. 
This supports our key insight: an unposed, background-free object image,
registered rather than supplied as a calibrated view, can recover much of the
benefit of a ground-truth second view with known pose.
Appearance-mismatched auxiliaries and a multi-object example are shown in
Appendix~\ref{sec:supp_neutral_ablation} and~\ref{sec:supp_multi_object}.

\begin{figure*}[t]
    \centering
    \includegraphics[width=0.8\linewidth]{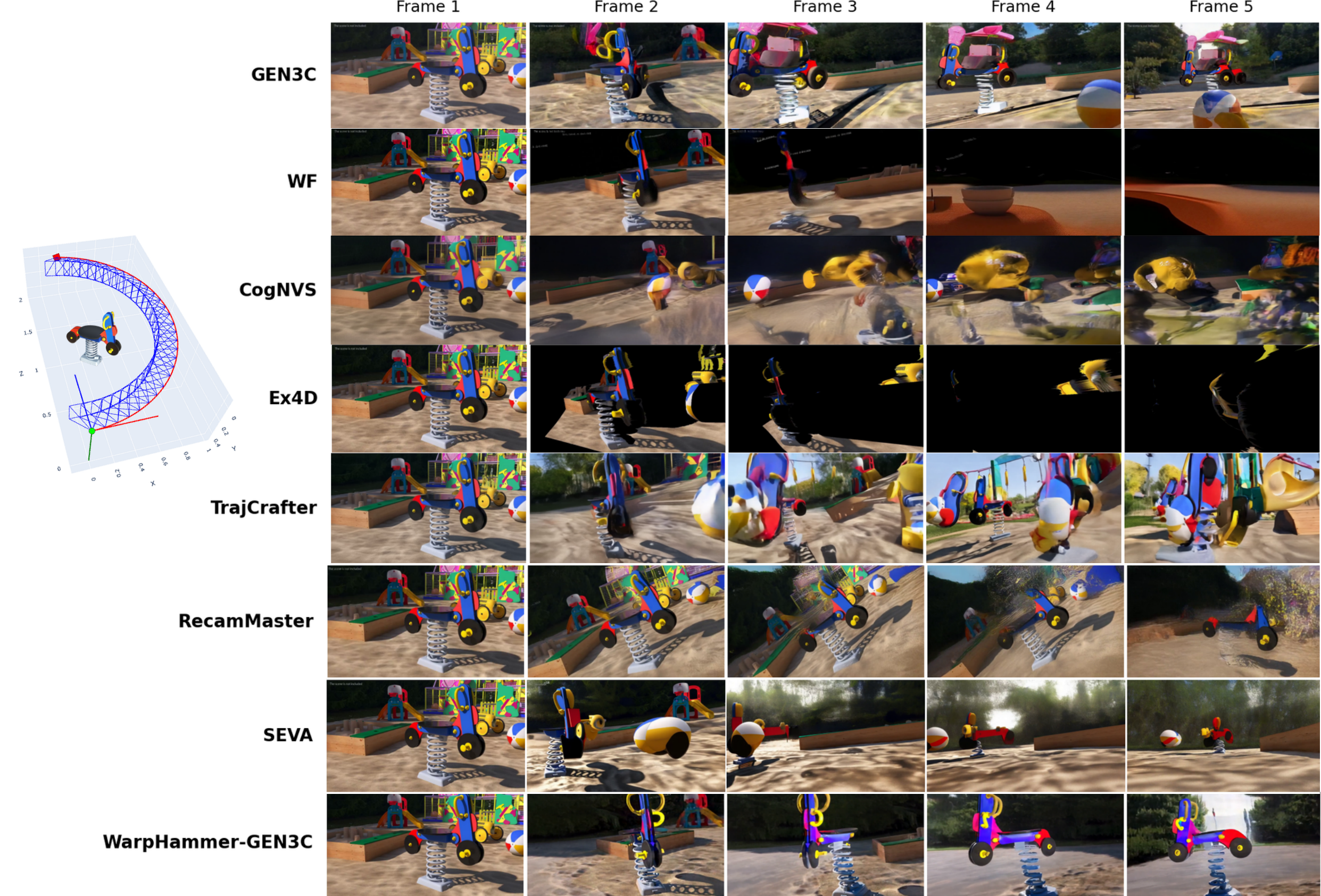}
    \caption{%
    \textbf{Qualitative comparison along an extreme orbital trajectory.}
    Single view NVS. Left: the target camera follows a wide arc around the central object (red curve). Right: five frames sampled from small to extreme orbital offsets.%
    }
    \label{fig:qual_main}
\end{figure*}

\begin{table*}[!htbp]
\centering
\caption{\textbf{Single-view} mean NVS results across 5 datasets (CO3D, Objectron, 360\textdegree{}, OmniObject3D, SynView-X). \ours{} variants are \colorbox{gray!10}{shaded}. Best results in \textbf{bold}. {\color{ForestGreen}Green} superscripts give differences from each backbone. $^{\dagger}$OrbErr is reported only on SynView-X.}
\label{tab:nvs_mean_sv_main}
\adjustbox{max width=\linewidth}{%
\begin{tabular}{l l ccc c ccc}
\toprule
& & \multicolumn{3}{c}{Image Similarity} & \multicolumn{1}{c}{Semantic} & \multicolumn{3}{c}{Camera Pose} \\
\cmidrule(lr){3-5} \cmidrule(lr){6-6} \cmidrule(lr){7-9}
Group & Method
& PSNR$\uparrow$ & SSIM$\uparrow$ & LPIPS$\downarrow$
& CLIP-F$\uparrow$
& RotErr$\downarrow$ & TransErr$\downarrow$ & OrbErr$^{\dagger}\downarrow$\\
\midrule
Camera-cond. & ReCamM.      & 7.91  & 0.481 & 0.851 & 93.42      & 163.49 & 0.102 & 0.3460 \\
Camera-cond. & SEVA         & 7.68  & 0.480 & 0.736 & 95.10      & 167.28 & 0.047 & 0.2855 \\
\midrule
Warp-based   & EX-4D        & 7.88  & 0.464 & 0.790 & 95.18      & 190.15 & 0.083 & 0.3192 \\
Warp-based   & CogNVS       & 8.08  & 0.513 & 0.768 & 94.26      & 176.51 & 0.063 & 0.3063 \\
Warp-based   & WF           & 7.64  & 0.524 & 0.568 & \textbf{95.73} & 119.85 & 0.085 & 0.3067 \\
Warp-based   & TrajCr.      & 7.58  & 0.460 & 0.791 & 93.62      & 177.85 & 0.069 & 0.3022 \\
Warp-based   & GEN3C        & 7.45  & 0.416 & 0.767 & 94.40      & 184.10 & 0.093 & 0.3058 \\
\midrule
\rowcolor{gray!10} Ours & \ours-TC     & 10.79\dg{+3.21} & 0.522 & 0.510 & 93.25 & 125.52\dg{-52.33} & 0.077 & 0.2288 \\
\rowcolor{gray!10} Ours & \ours-GEN3C  & \textbf{11.67}\dg{+4.22} & \textbf{0.589} & \textbf{0.444} & 94.34 & \textbf{91.21}\dg{-92.89} & \textbf{0.024} & \textbf{0.2045} \\
\bottomrule
\end{tabular}
}
\end{table*}

\begin{table*}[!htbp]
\centering
\caption{{\bf Auxiliary-view} NVS results averaged across 5 datasets (CO3D, Objectron, 360\textdegree{}, OmniObject3D, SynView-X). \ours{} is the \colorbox{gray!10}{shaded row}. \textcolor{gray}{Gray} rows receive GT poses and in-scene second views and are stronger-supervision references, not equal-input baselines. \textbf{Bold}: best among non-gray rows.}
\label{tab:nvs_mean_mv_main}
\adjustbox{max width=\linewidth}{%
\begin{tabular}{l l ccc c cc}
\toprule
& & \multicolumn{3}{c}{Image Similarity} & \multicolumn{1}{c}{Semantic} & \multicolumn{2}{c}{Camera Pose} \\
\cmidrule(lr){3-5} \cmidrule(lr){6-6} \cmidrule(lr){7-8}
Setting & Method
& PSNR$\uparrow$ & SSIM$\uparrow$ & LPIPS$\downarrow$
& CLIP-F$\uparrow$
& RotErr$\downarrow$ & TransErr$\downarrow$ \\
\midrule
\textcolor{gray}{GT poses} & \textcolor{gray}{SEVA-MV}  & \textcolor{gray}{14.60} & \textcolor{gray}{0.705} & \textcolor{gray}{0.389} & \textcolor{gray}{94.61} & \textcolor{gray}{193.04} & \textcolor{gray}{0.040} \\
\textcolor{gray}{GT poses} & \textcolor{gray}{GEN3C-MV} & \textcolor{gray}{14.31} & \textcolor{gray}{0.696} & \textcolor{gray}{0.446} & \textcolor{gray}{93.06} & \textcolor{gray}{176.96} & \textcolor{gray}{0.069} \\
\midrule
No GT poses & SEVA-MV   & 10.56 & 0.513 & 0.519 & 92.89 & 97.13 & 0.050 \\
No GT poses & GEN3C-MV  & 9.97  & 0.502 & 0.546 & 93.30 & 81.38 & 0.088 \\
\midrule
\rowcolor{gray!10} No GT poses & \ours-GEN3C & \textbf{14.09} & \textbf{0.635} & \textbf{0.389} & \textbf{94.84} & \textbf{80.03} & \textbf{0.035} \\
\bottomrule
\end{tabular}
}
\end{table*}

\paragraph{Qualitative.}
Fig.~\ref{fig:qual_main} shows a qualitative comparison along 
a wide orbital trajectory. Warp-based baselines preserve the 
input view at small offsets but progressively lose the object 
as the warp collapses: mirror-like duplications and background 
leakage dominate at wide angles. Camera-conditioned generation 
(ReCamMaster) avoids warping artifacts but fails to follow the 
prescribed orbital trajectory, producing a near-static sequence that barely 
changes viewpoint. \ours{}-GEN3C maintains a coherent, identity-preserving 
rendering across the full orbital range, follows the prescribed 
camera trajectory, and produces plausible background completion 
in newly visible regions. More qualitative results and animated comparisons are provided in Appendix~\ref{sec:supp_qual}.

\FloatBarrier

\subsection{Ablation}
\label{sec:ablation}
We ablate the two core components of \ours{} on 50 SynView-X samples: 
the object-centric 3D prior and the VGGT-based multi-view fusion. 
We study four variants across two settings: \textit{Single-view Full} 
vs.\ \textit{w/o 3D prior}, and \textit{Auxiliary-view Full} vs.\ 
\textit{w/o VGGT}. Results are reported in 
Tab.~\ref{tab:ablation_synviewx_main}.

\paragraph{Effect of the 3D prior (single-view).}
Removing the explicit 3D object reconstruction 
(\textit{Single-view w/o 3D prior}) sharply degrades both image 
fidelity and camera accuracy simultaneously, confirming that the 3D 
prior serves a dual role: it supplies the missing surface geometry for 
newly visible object regions, and restores the implicit camera cue that 
the warp normally carries but loses when it goes sparse at extreme angles.

\paragraph{Effect of auxiliary-view fusion (VGGT).}
Comparing \textit{Aux-view Full} against \textit{Aux-view w/o VGGT} 
isolates the contribution of VGGT-based geometric alignment between the 
auxiliary view and the reference scene. Even without VGGT, SAM3D still 
produces a plausible object shape; what VGGT provides is the joint alignment that places the 
auxiliary geometry consistently inside the scene coordinate frame. 
Without it, the orbital trajectory drifts relative to the object, confirming that aligned conditioning,
not 3D content alone, is the critical ingredient for orbital fidelity.

\begin{table}[!htbp]
\centering
\caption{Ablation study on SynView-X (n=50).  Best results per setting in \textbf{bold}.}
\label{tab:ablation_synviewx_main}
\footnotesize
\setlength{\tabcolsep}{5pt}
\renewcommand{\arraystretch}{1.25}
\begin{tabular}{l l ccc c ccc}
\toprule
& & \multicolumn{3}{c}{Image Similarity} & \multicolumn{1}{c}{Semantic} & \multicolumn{3}{c}{Camera Pose} \\
\cmidrule(lr){3-5} \cmidrule(lr){6-6} \cmidrule(lr){7-9}
Setting & Variant
& PSNR$\uparrow$ & SSIM$\uparrow$ & LPIPS$\downarrow$
& CLIP-F$\uparrow$
& RotErr$\downarrow$ & TransErr$\downarrow$ & OrbErr$\downarrow$ \\
\midrule
Aux-view & Full        & \B{16.69} & \B{0.560} & \B{0.368} & \B{94.94} & \B{25.36} & \B{0.134} & \B{0.132} \\
Aux-view & w/o VGGT    & 9.80  & 0.360 & 0.609 & 94.83 & 49.37 & 0.154 & 0.192\\
\midrule
Single-view & Full        & \B{12.06} & \B{0.468} & \B{0.482} & 94.50 & \B{31.05} & \B{0.132} & \B{0.1338} \\
Single-view & w/o 3D prior& 10.05 & 0.360 & 0.600 & \B{94.84} & 49.45 & 0.154 & 0.1772 \\
\bottomrule
\end{tabular}
\end{table}

Original-RGB and appearance-neutralized auxiliary conditioning are compared in
\Cref{fig:ablation_donor_mismatch} in the Appendix.
\FloatBarrier

\section{Summary and Conclusions}
\ours{} addresses a key limitation of projection-conditioned NVS: under large viewpoint changes, the scene warp becomes sparse around the foreground object, weakening both the geometric support and the implicit camera cue provided to the generator. We overcome this by densifying the warped conditioning signal with an explicit object-centric 3D reconstruction.
The resulting representation also opens a practical new use case for NVS: incorporating an unposed, out-of-scene auxiliary image of the object. Across synthetic, scanned, and real-world benchmarks, our results show that object-centric geometry completion is an effective path toward more robust, flexible, and practically usable NVS systems.

\paragraph{Limitations.}
\label{sec:limitations}

\ours{} relies on accurate object segmentation and the quality of the
external 3D reconstruction; segmentation failures or incorrect geometry
propagate into the synthesized views (Appendix~\ref{sec:supp_mask_robustness}). The method is most effective for
scenes with a dominant foreground object and may be less reliable for
cluttered scenes, thin structures, or transparent materials. The
auxiliary-view setting is sensitive to geometric ambiguities when the
auxiliary and reference views are either nearly identical or highly dissimilar.
Finally, adding SAM3D to the pipeline increases inference time relative
to the base backbone, which may be a consideration in latency-sensitive
applications.

\clearpage

\bibliography{iclr2027_conference}
\bibliographystyle{iclr2027_conference}

\clearpage

\appendix

\begin{center}
{\Large\bfseries Supplementary Material:\\WarpHammer: Densifying Scene Warps with 3D \\ Object Priors for Extreme View Synthesis}\\[1ex]
\end{center}
\vspace{1em}

In this supplementary material, we provide a multi-object example, implementation
and dataset details, and additional quantitative and qualitative results.

\section{Multi-Object View Synthesis}
\label{sec:supp_multi_object}

\begin{figure}[!htbp]
\centering
\begin{tabular}{cc}
Reference & Supplied auxiliary\\
\includegraphics[width=0.46\linewidth]{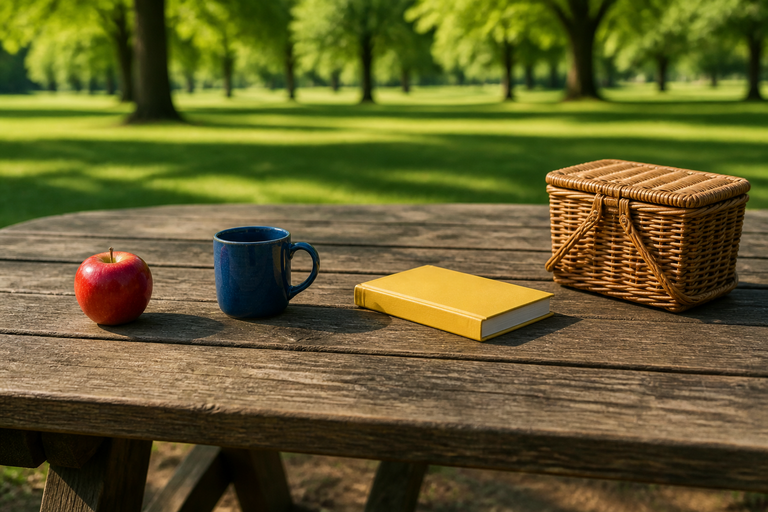} &
\includegraphics[width=0.46\linewidth]{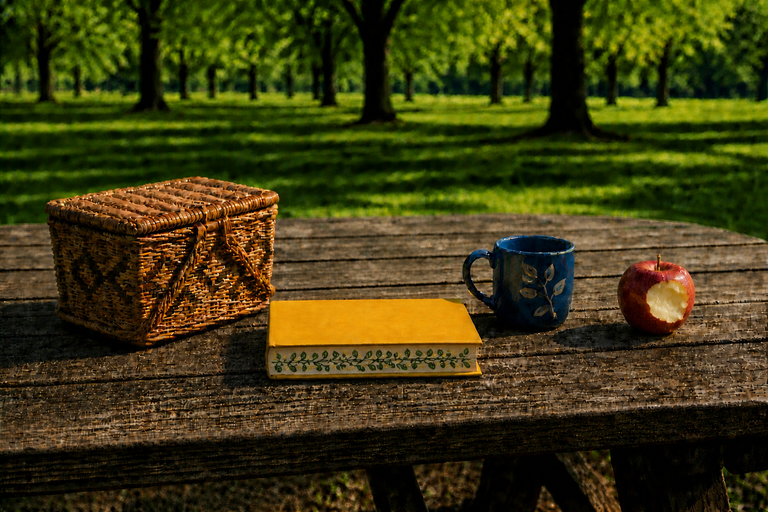}
\end{tabular}
\par\smallskip
{\small Frames 0, 30, 60, 90 and 120}\\
{\small Composed warp}\\
\includegraphics[width=\linewidth]{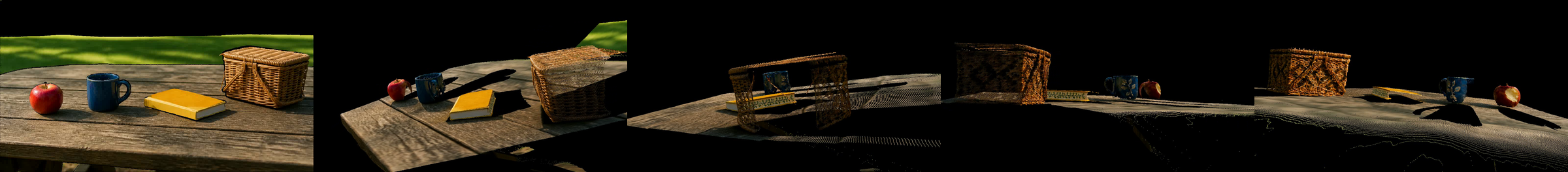}\\
{\small Diffusion output}\\
\includegraphics[width=\linewidth]{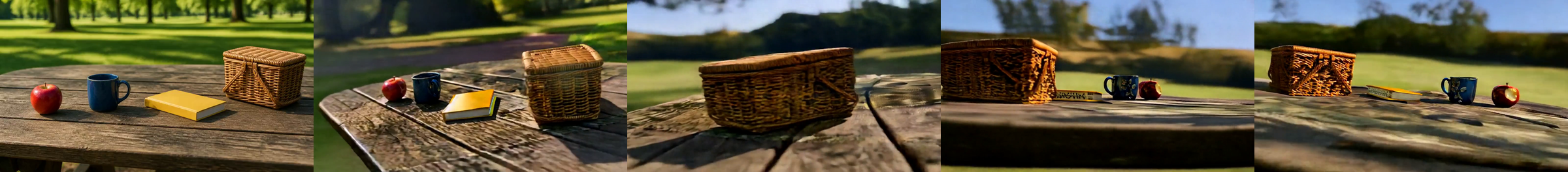}
\caption{Multi-object synthesis over a $200^\circ$ orbit. The regenerated
auxiliary supplies additional object views.}
\label{fig:multi_picnic2}
\end{figure}

The object-centric representation extends naturally to scenes with multiple
foreground objects. We reconstruct each object from its reference and auxiliary
crops, place its two-view geometry into the scene with one similarity transform,
and render the composed cache to condition video generation.
The auxiliary image is not a view of the target scene: it shows the same
objects in a different arrangement, each seen from a different viewpoint. Since
each object is registered with its own transform, the auxiliary does not
correspond to any single camera of the target scene.
Note that the auxiliary image is not used as a whole: only the segmented
crops of the individual objects are used, so its background, table and object
arrangement provide no information about the target scene.
\Cref{fig:multi_picnic2} illustrates a $200^\circ$ orbit around a picnic scene
containing an apple, mug, book and basket. The full video is provided on the
project page.

\FloatBarrier

\section{Implementation Details}
\label{sec:supp_impl}

\subsection{Choice of 3D Generator}
\label{sec:supp_generator}

We instantiate the 3D generator $f_{3D}$ with SAM3D~\citep{sam3d}, which accepts the masked reference point cloud $P_r$ as conditioning and returns the object mesh directly in the reference camera frame and at the scale of $P_r$. Other image-to-3D generators, such as TRELLIS~\citep{xiang2025structured} and Hunyuan3D-2~\citep{zhao2025hunyuan3d}, can also be used as $f_{3D}$. Since these generators reconstruct objects in a canonical frame, their output requires an additional registration step that aligns it with $P_r$ before it is fused into the cache.

 \subsection{Cross-View Point-Map Alignment}
 \label{sec:supp_alignment}
 
 The similarity transform in \Cref{eq:aux_align} maps the VGGT prediction into the frame and scale of the reference cache. Because $P_r$ and $P_r^g$ are indexed by the same reference pixels, they provide direct 3D correspondences on the set $\mathcal{U}$ of object pixels valid in both. The transform is computed in closed form. We first center both point sets at their medians over $\mathcal{U}$, $\bar{P}_r$ and $\bar{P}_r^g$; $s$ is the ratio of their median distances to these centers, $\mathbf{R}$ is the least-squares rotation between the centered, rescaled correspondences, obtained from a singular value decomposition with the Kabsch algorithm~\citep{kabsch1976solution}, and $\mathbf{t}=\bar{P}_r-s\mathbf{R}\bar{P}_r^g$ aligns the centers. Medians make the fit robust to outlier points. Before alignment, confidence filtering discards VGGT points below the $25$th percentile of the per-view confidence.

\subsection{Appearance-Neutral Auxiliary Conditioning}
\label{sec:supp_aux_neutral}

We use depth-coded auxiliary conditioning to incorporate geometric support
without copying mismatched donor color. We refer to it as appearance-neutral
conditioning, and to replacing auxiliary RGB with it as neutralization. For appearance-matched captures, as in
the main results, we retain auxiliary RGB: $\tilde{R}^a_t$ is the auxiliary
rendering itself.
The auxiliary footprint is obtained automatically by projecting the aligned
auxiliary geometry into each target view; it determines where the auxiliary
conditioning is applied. A configuration flag enables neutralization when
needed; auxiliary RGB is retained otherwise.

\paragraph{Motivation.}
An external object image can provide useful geometry despite differences in
color, texture or illumination. Neutralization separates this geometric
information from the donor's RGB appearance, allowing the reference image to
guide the synthesized appearance. \Cref{sec:supp_neutral_ablation} compares
RGB and depth-coded conditioning under matched and mismatched donor colors.

\paragraph{Depth-coded buffer and confidence mask.}
Let $R^a_t$ denote the auxiliary RGB rendering at target frame $t$, $D^a_t$ the projected auxiliary depth cue, $R^s_t$ the rendering from the reference scene cache, and $\Omega^a_t$ the rendered auxiliary object footprint. We define the depth-coded auxiliary buffer $\tilde{R}^a_t$ and mask $M^a_t$ jointly as
\begin{equation}
\big(\tilde{R}^a_t(u,v),\; M^a_t(u,v)\big)
=
\begin{cases}
\big(g(D^a_t(u,v)),\; \alpha\big),
& (u,v)\in\Omega^a_t,\\[2pt]
\big(R^s_t(u,v),\; 0\big),
& \text{otherwise.}
\end{cases}
\label{eq:neutral_joint}
\end{equation}
Here, $g(\cdot)$ converts the auxiliary depth rendering into an appearance-neutral RGB conditioning signal, and $\alpha\in(0,1)$ is a constant confidence inside the auxiliary footprint, used in place of the validity mask when the donor color is not to be trusted. Setting $\alpha<1$ lets the prior use the depth code as a structural hint rather than reproduce it as gray appearance. When auxiliary RGB is retained, the renderer's validity mask is used instead.

\paragraph{Implementation.}
We normalize auxiliary depth using the valid depth range at cue construction:
\begin{equation}
\bar{D}^a_t(u,v)
=
\frac{D^a_t(u,v)-d_{\min}}
{d_{\max}-d_{\min}+\epsilon},
\end{equation}
where $d_{\min}$ and $d_{\max}$ are the minimum and maximum valid depths and
$\epsilon$ ensures numerical stability. In the cache-based car experiments,
normalization is applied before projection; the normalized depth attributes
are then rendered along the trajectory. The depth-coding function $g$
repeats the normalized scalar depth value across the RGB channels:
\begin{equation}
g(D^a_t(u,v))
=
\big(\bar{D}^a_t(u,v),\bar{D}^a_t(u,v),\bar{D}^a_t(u,v)\big).
\label{eq:neutral_g}
\end{equation}
Within the auxiliary footprint $\Omega^a_t$, the depth-coded
rendering replaces the auxiliary RGB contribution in GEN3C's warped conditioning,
with $M^a_t$ providing the corresponding confidence mask.

\begin{figure*}[!htbp]
    \centering
    \begin{tabular}{cc}
    Reference & Colored auxiliary\\
    \includegraphics[width=0.30\linewidth]{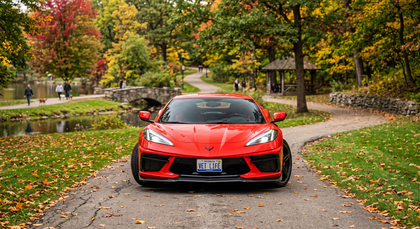} &
    \includegraphics[width=0.30\linewidth]{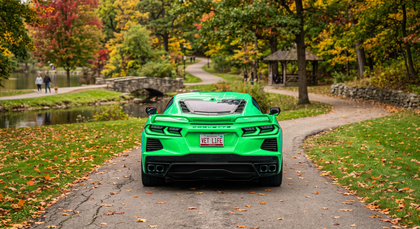}
    \end{tabular}\\
    {\small RGB-conditioned output}\\
    \includegraphics[width=\linewidth]{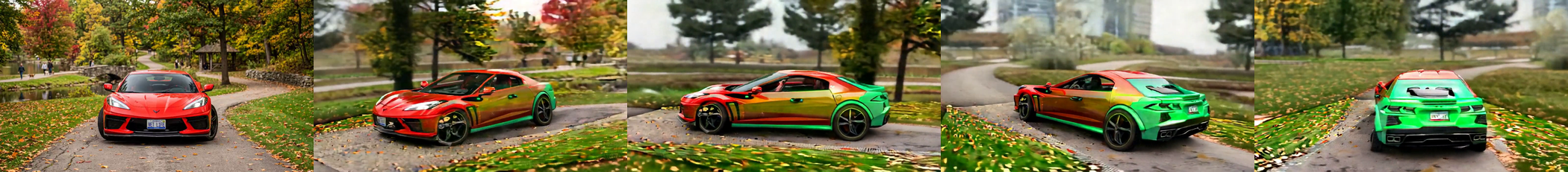}\\
    {\small Grayscale-conditioned output}\\
    \includegraphics[width=\linewidth]{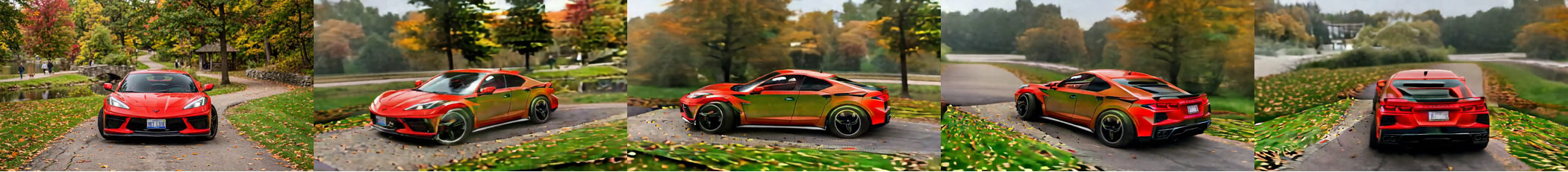}
    \caption{Grayscale conditioning in auxiliary-view synthesis.
    Above: reference and hue-rotated auxiliary. Rows show RGB-conditioned
    synthesis with the colored auxiliary and the grayscale-conditioned result,
    obtained from the luminance of the original auxiliary.}
    \label{fig:ablation_donor_mismatch}
\end{figure*}

Depth coding and luminance conversion are distinct: the former replaces RGB
with a geometric signal, whereas the latter repeats the grayscale luminance of $R^a_t$ across RGB
channels, removing chromatic color while retaining intensity variation.
The grayscale-conditioned example in \Cref{fig:ablation_donor_mismatch}
uses the latter.

\subsection{Ablation of Appearance-Neutral Conditioning}
\label{sec:supp_neutral_ablation}

\paragraph{Suppressing donor-color transfer.}
\Cref{fig:qual_website_fusion} shows
two cross-instance examples pairing an orange reference car with a pink
auxiliary car. These qualitative results illustrate appearance preservation
with a different donor instance; the project page provides
the full videos.

\subsection{Segmentation Sensitivity}
\label{sec:supp_mask_robustness}

\begin{figure*}[!htbp]
    \centering
    \includegraphics[width=\linewidth]{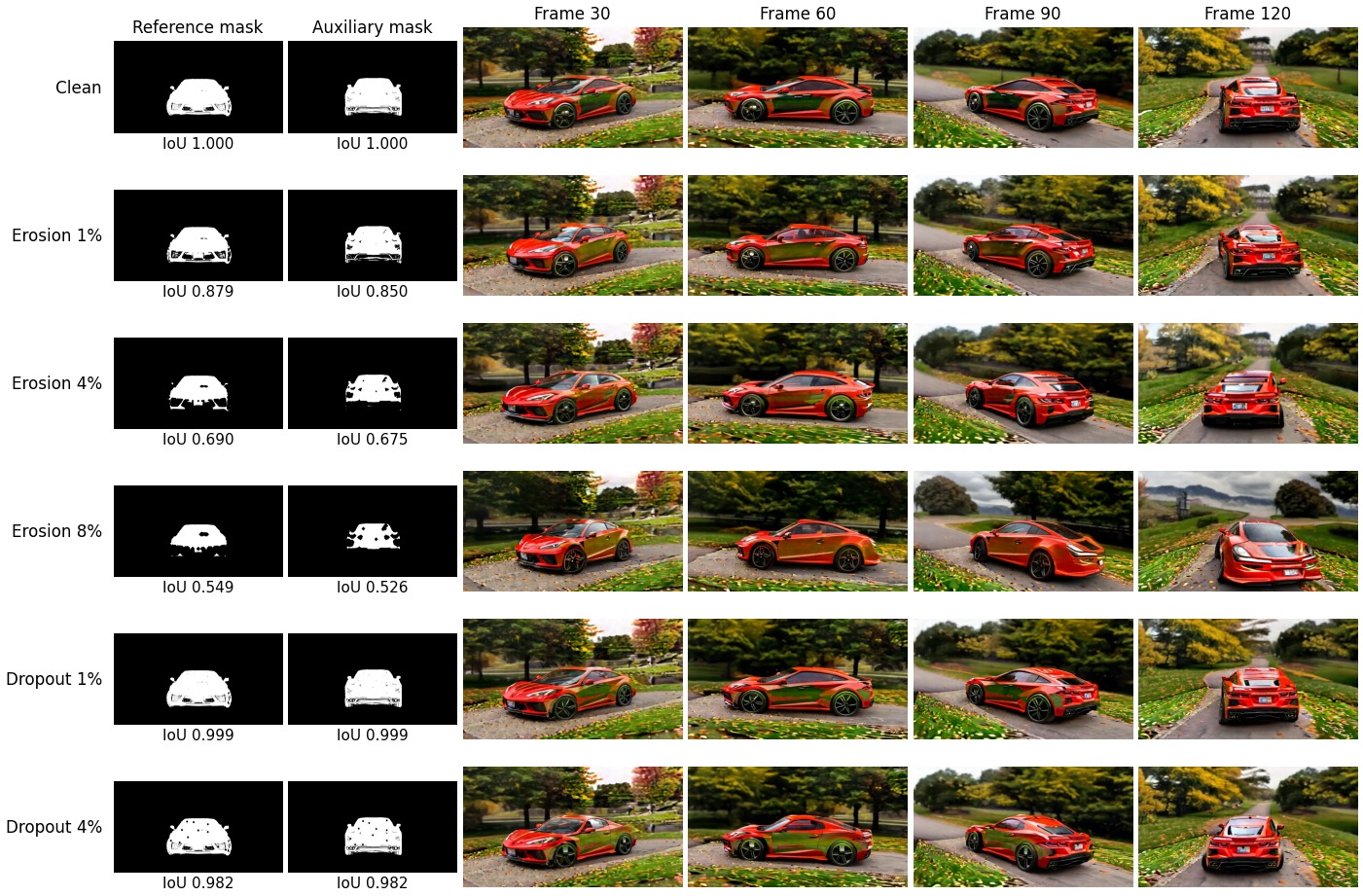}
    \caption{Segmentation sensitivity on a fixed-seed local example.
    Left: reference and auxiliary masks, with IoU relative to
    their clean counterparts. Right: corresponding output frames.
    Percentages specify perturbation radius, not mask-area change.
    Erosion and dropout up to $4\%$ preserve a recognizable vehicle, while
    $8\%$ erosion alters the body.}
    \label{fig:mask_robustness}
\end{figure*}

We ablate the sensitivity of our method to errors in the segmentation masks.
We evaluate 6 conditions on one front/rear pair with fixed perturbation and
generation seeds: clean masks, erosion at radii of $1\%,4\%,8\%$ and interior
dropout at radii of $1\%,4\%$ of the object's equivalent radius $\sqrt{A/\pi}$.
Reconstruction and synthesis are rerun for each condition with the same
preprocessing. Erosion and dropout up to $4\%$ retain a recognizable single
vehicle throughout the trajectory, despite changes in appearance and
proportions, while $8\%$ erosion strongly alters the body
(\Cref{fig:mask_robustness}).

\FloatBarrier
\section{Dataset Details}
\label{sec:supp_datasets}

\subsection{Dataset Creation}
\label{sec:supp_datasets_creation}

For each benchmark, we construct evaluation pairs by selecting two views of the same object (or scene) separated by a large orbital offset. The first view serves as the {\it reference} input, the second as the auxiliary target, and a sequence of intermediate frames provides the ground-truth orbital video. All camera poses are converted to a common OpenCV world-to-camera convention. The dataset-specific selection and pre-processing protocols are described below.

\paragraph{CO3D~\citep{reizenstein2021common} (160 pairs).}
For each sequence, we greedily select up to $20$ pairs that maximize frame separation while satisfying $|\text{yaw}|\!\in\![100^{\circ},190^{\circ}]$ and $|\text{pitch}|\!\le\!40^{\circ}$. Pairs are further filtered by foreground-mask coverage in $[5\%,95\%]$ using the segmentation masks provided with CO3D. The ground-truth orbital video consists of the ordered real frames lying between the two selected views.

\paragraph{Mip-NeRF~360~\citep{mipnerf360} (57 pairs).}
Camera poses are taken from the COLMAP reconstructions provided with the dataset, and the dense scene point cloud is used to estimate the 3D scene centre, which serves as the per-view crop anchor. Foreground masks are extracted by an off-the-shelf saliency segmentation model and made cross-view consistent by projecting the foreground 3D points from one view into the other and retaining only mask components that overlap with these projections. Intermediate ground-truth frames are obtained by traversing the shorter azimuthal arc between the two selected cameras.

\paragraph{OmniObject3D~\citep{wu2023omniobject3d} (500 pairs).}
We use the rendered images and camera poses provided by OmniObject3D. For each object, a pair satisfying our orbital-rotation criteria is sampled at random from all qualifying combinations. The reference view is composited onto a randomly sampled perspective crop from an HDRI panorama to simulate a realistic background, while the auxiliary view is placed on a white background, yielding the unposed object input used by \ours{}. Ground-truth video frames are taken as the rendered frames closest, in spherical distance, to a linearly interpolated arc between the two selected views.

\paragraph{Objectron~\citep{ahmadyan2021objectron} (9 pairs).}
We decode the original sequences and convert their camera transforms to our common OpenCV convention. Per sequence, we select the pair with the largest in-range $|\text{yaw}|$, prioritizing pairs whose annotated 3D bounding box is well-framed. Foreground masks are obtained from an off-the-shelf saliency model, and intermediate frames are taken directly from the source monocular videos.

\paragraph{SynView-X (900 pairs).}
SynView-X aggregates synthetic and scanned assets rendered to support extreme orbital viewpoint changes. It is composed of $777$ pairs from ShapeNet\-Core~\citep{5shapenet}, $20$ pairs from GSO~\citep{gso}, $103$ pairs from Objaverse~\citep{objaverse}. The ShapeNet, GSO, and Objaverse assets are rendered offline; for ShapeNet and GSO, yaw and pitch are sampled deterministically per asset to ensure reproducibility, with the orbit direction matching the conditioning convention used by GEN3C so that the stored angles can be passed directly as conditioning signals. Together, these subsets cover man-made objects, real-scanned assets, and procedurally generated scenes, giving SynView-X the diversity needed to stress-test extreme-viewpoint synthesis under controlled conditions.

\paragraph{Pair-selection summary.}
Across all five benchmarks, evaluation pairs lie on an orbital arc around the object of interest, with yaw separation in $[100^{\circ},190^{\circ}]$ and bounded pitch. The reference view is provided as input to all methods, the auxiliary view is used as the unposed conditioning input for \ours{} auxiliary-view variants. In the auxiliary-view setting, endpoint image similarity is therefore measured at the view whose masked object is also supplied, without pose, as input; its pose, background and all intermediate frames are held out. Intermediate frames define the ground-truth orbital video for video-quality and OrbErr computation.
\section{Evaluation Metrics Details}
\label{sec:supp_metrics}

\subsection{Orbital Evaluation: Limitations of RotErr and Definition of OrbErr}
\label{sec:supp_orberr}

The standard relative rotation error
\begin{equation}
E_{\mathrm{rot}}(t) =
\cos^{-1}\!\left(
\frac{\operatorname{tr}(\hat{R}_t R_t^\top)-1}{2}
\right)
\label{eq:roterr_geodesic}
\end{equation}
measures the geodesic distance on $\mathrm{SO}(3)$ between the predicted and ground-truth rotation matrices for frame $t$. Following CameraCtrl~\citep{he2025cameractrl} and ReCamMaster~\citep{bai2025recammaster}, we compute this per-frame error and sum it over all frames of a sequence,
\begin{equation}
\mathrm{RotErr} = \sum_{t=1}^{T} E_{\mathrm{rot}}(t),
\label{eq:roterr_sum}
\end{equation}
and report the mean sequence-level RotErr across evaluation samples, in degrees. Because \Cref{eq:roterr_sum} accumulates over $T$ frames rather than averaging, the reported values are sequence totals and are not bounded by the $180^\circ$ maximum of a single-frame geodesic distance; the same accumulation convention is applied to every method, so the comparison remains like-for-like. This is a useful metric for general camera-control evaluation, but it is not fully aligned with object-centric orbital novel view synthesis. In our setting, the desired motion is primarily defined by the orbital viewpoint around the object, i.e., its azimuth and elevation, rather than by an arbitrary 3D rotation.

\paragraph{TransErr.}
Following CameraCtrl~\citep{he2025cameractrl}, we measure the translation error as the summed squared $\ell_2$ distance between the ground-truth and estimated camera translations,
\begin{equation}
\mathrm{TransErr} = \sum_{t=1}^{T} \big\lVert T_t - \hat{T}_t \big\rVert_2^2,
\label{eq:transerr}
\end{equation}
where $T_t$ and $\hat{T}_t$ are the ground-truth and estimated translation vectors at frame $t$. Because monocular pose estimates are defined only up to scale, both trajectories are first expressed relative to their first frame, and the estimated trajectory is rescaled so that its translation between the first two frames matches that of the ground truth.

\paragraph{Conflation of orbital and non-orbital errors.}
RotErr collapses all rotational degrees of freedom into a single scalar and does not distinguish whether the error originates in azimuth, elevation, or in-plane roll. Azimuth and elevation determine the object-centric viewpoint, while roll changes the image-plane orientation without necessarily changing the orbital position around the object. Concretely, consider two predictions with the same geodesic rotation error:

1. $\hat{R}_t^{(1)}$: azimuth error of $30^\circ$, with no elevation or roll error. 2. $\hat{R}_t^{(2)}$: roll error of $30^\circ$, with no azimuth or elevation error.

\begin{figure*}[!htbp]
    \centering
    \includegraphics[width=\linewidth]{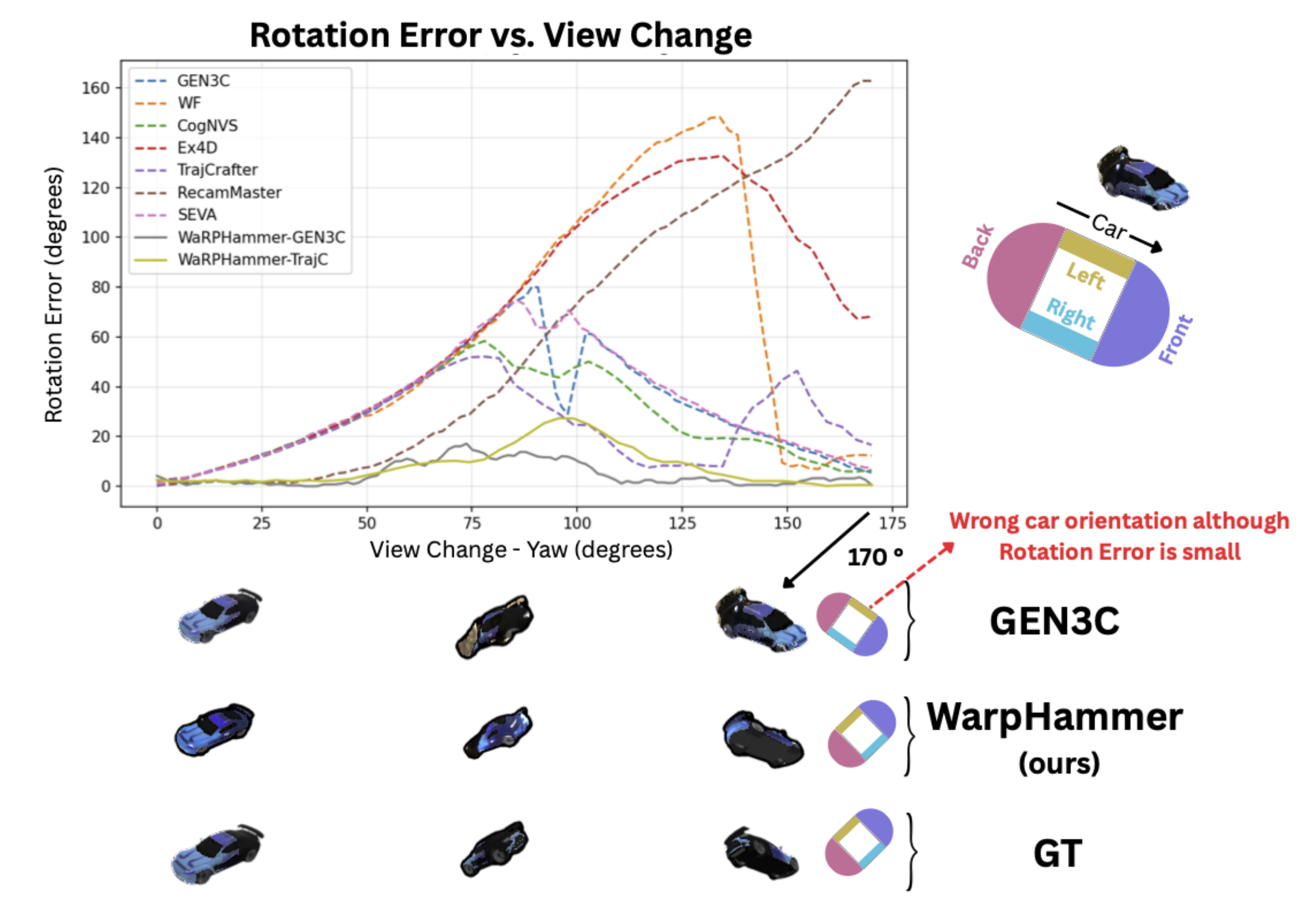}
    \caption{\textbf{Rotation error vs. view change.} Despite decreasing rotation error at large viewpoint changes, the predicted object orientation can still be incorrect (see comparison with GT), indicating that rotation error alone is insufficient for evaluating object-centric consistency in orbital settings.}
    \label{fig:rot_error_view_change}
\end{figure*}

Both yield $E_{\mathrm{rot}}(t)=30^\circ$, yet they represent different failure modes: the first observes the object from the wrong orbital viewpoint, while the second preserves the orbital direction but introduces an in-plane rotation. For object-centric orbital synthesis, these errors should not be treated as equivalent. In Figure~\ref{fig:rot_error_view_change}, we illustrate such a failure case: GEN3C achieves a low rotation error under large viewpoint changes, yet the car orientation is incorrect, indicating that the metric fails to capture this error.

\paragraph{Insensitivity to orbital position.}
RotErr depends only on rotation matrices and ignores camera translation entirely. Even if rotation is perfectly recovered, a camera centre that lies off the intended orbital arc is not penalized. Conversely, under monocular pose estimation, absolute translation is unrecoverable up to scale, so naively reporting translation error is unreliable. A metric tailored to orbital fidelity must therefore evaluate the orbital viewpoint without depending on metric translation.

\paragraph{OrbErr.}
We introduce OrbErr, a metric designed for object-centric orbital evaluation. OrbErr combines two complementary terms: (i)~a rotation-derived Euler-angle error that measures azimuth and elevation agreement, and (ii)~a feature-displacement error that checks whether object features appear at their expected spatial positions in the generated frame. This allows the metric to penalize cases where the camera estimate appears plausible but the object is rendered from an incorrect orbital viewpoint.

Given monocular pose estimates $\{\hat{R}_t\}_{t=0}^{T}$ and target rotations $\{R_t\}_{t=0}^{T}$, where $t=0$ denotes the reference view and $t=1,\dots,T$ the generated frames of \Cref{sec:method_prelim}, we first compute rotations relative to the initial frame:
\begin{equation}
\hat{R}_t^{\mathrm{rel}} = \hat{R}_t \hat{R}_0^\top,
\qquad
R_t^{\mathrm{rel}} = R_t R_0^\top .
\end{equation}
We extract the camera forward direction as the third row of the relative rotation,
\begin{equation}
\hat{f}_t = e_3^\top \hat{R}_t^{\mathrm{rel}},
\qquad
f_t = e_3^\top R_t^{\mathrm{rel}},
\qquad
\hat{f}_t, f_t \in \mathbb{R}^3 ,
\end{equation}
and decompose it into azimuth and elevation:
\begin{equation}
\hat{\theta}_t = \operatorname{atan2}(\hat{f}_{t,x}, \hat{f}_{t,z}),
\qquad
\theta_t = \operatorname{atan2}(f_{t,x}, f_{t,z}),
\end{equation}
\begin{equation}
\hat{\phi}_t =
\operatorname{atan2}\!\left(-\hat{f}_{t,y},
\sqrt{\hat{f}_{t,x}^2+\hat{f}_{t,z}^2}\right),
\qquad
\phi_t =
\operatorname{atan2}\!\left(-f_{t,y},
\sqrt{f_{t,x}^2+f_{t,z}^2}\right).
\end{equation}
The wrapped angular distance is
\begin{equation}
d_{\angle}(a,b)
=
\left|
\operatorname{atan2}\!\left(\sin(a-b),\cos(a-b)\right)
\right|.
\end{equation}
We define the normalized Euler-angle orbital error as
\begin{equation}
E_{\mathrm{euler}}(t)
=
\frac{1}{2}
\frac{d_{\angle}(\hat{\theta}_t,\theta_t)}{\pi}
+
\frac{1}{2}
\frac{d_{\angle}(\hat{\phi}_t,\phi_t)}{\pi}.
\end{equation}
This term measures whether the generated frame corresponds to the intended orbital viewing direction, focusing on azimuth and elevation rather than arbitrary 3D rotation error.

To additionally verify that the rendered object appearance agrees with the intended viewpoint, we compute a feature-displacement error using DINOv2~\citep{oquab2023dinov2} patch features on a $G \times G$ grid, restricted to object masks obtained from SAM~2~\citep{ravi2024sam}. Unlike a one-sided nearest-neighbor distance, which can match many ground-truth patches to the same generated patch, we use mutual nearest-neighbor matches between the ground-truth and generated object regions. Specifically, nearest neighbors are 
computed in both directions between the generated and ground-truth object 
patches, and only reciprocal matches are retained.

Let $\Omega_t^{\mathrm{gt}}$ and $\Omega_t^{\mathrm{gen}}$ denote the valid object patches in the ground-truth and generated frames, respectively. We $\ell_2$-normalize all DINOv2 patch features and compute nearest neighbors by cosine similarity. For each ground-truth patch $i\in\Omega_t^{\mathrm{gt}}$, let $\operatorname{NN}_{\mathrm{gen}}(i)$ be its nearest generated patch. Similarly, for each generated patch $j\in\Omega_t^{\mathrm{gen}}$, let $\operatorname{NN}_{\mathrm{gt}}(j)$ be its nearest ground-truth patch. The set of mutual nearest-neighbor matches is
\begin{equation}
\mathcal{M}_t
=
\left\{
(i,j):
j=\operatorname{NN}_{\mathrm{gen}}(i),
\;
i=\operatorname{NN}_{\mathrm{gt}}(j)
\right\}.
\end{equation}
The mutual feature-displacement error is
\begin{equation}
E_{\mathrm{disp}}^{\mathrm{mnn}}(t)
=
\frac{1}{|\mathcal{M}_t|}
\sum_{(i,j)\in\mathcal{M}_t}
\frac{\|p_i-p_j\|_2}{\sqrt{2}(G-1)},
\end{equation}
where $p_i$ and $p_j$ are the two-dimensional grid coordinates of the matched patches. The normalization by $\sqrt{2}(G-1)$ converts the displacement to units of the grid diagonal.

Because mutual matching may produce a small set of confident matches even when the generated object is incomplete, we also include a coverage penalty:
\begin{equation}
E_{\mathrm{cov}}(t)
=
1-
\frac{2|\mathcal{M}_t|}
{|\Omega_t^{\mathrm{gt}}|+|\Omega_t^{\mathrm{gen}}|}.
\label{eq:orberr_cov}
\end{equation}
The final feature-displacement term is
\begin{equation}
E_{\mathrm{disp}}(t)
=
\frac{1}{2}E_{\mathrm{disp}}^{\mathrm{mnn}}(t)
+
\frac{1}{2}E_{\mathrm{cov}}(t).
\end{equation}
If no mutual matches are found, we set $E_{\mathrm{disp}}(t)=1$.

The final orbital error averages both components:
\begin{equation}
E_{\mathrm{orb}}(t)
=
\frac{1}{2}E_{\mathrm{euler}}(t)
+
\frac{1}{2}E_{\mathrm{disp}}(t).
\end{equation}
OrbErr therefore evaluates both the estimated orbital viewing direction and the spatial consistency of object features. The mutual nearest-neighbor constraint makes the feature-displacement term more robust to many-to-one matches, repeated textures, and mirror-like duplications, while the coverage penalty discourages incomplete or weakly supported object reconstructions from receiving artificially low error. We use OrbErr as the primary measure of orbital fidelity, while still reporting RotErr for compatibility with prior camera-controlled video generation work.

\paragraph{Sanity check with known cameras.}
To verify that OrbErr responds to orbital and structural errors rather than to appearance changes, we implemented the full metric above and evaluated it on controlled inputs with exact cameras. We procedurally render three asymmetric, textured box assemblies (independent of our reconstructions) along a five-frame orbit, azimuth $0^\circ$ to $120^\circ$ in $30^\circ$ steps. Keeping the first frame fixed, we modify frames 2--5 in three ways: (i)~re-rendering them with a yaw, pitch, or roll offset of $5^\circ$, $15^\circ$, or $30^\circ$; (ii)~appearance edits (hue $+10^\circ$, brightness $+10\%$, Gaussian blur $\sigma{=}2$\,px); and (iii)~structural edits with unchanged cameras (removing a central band covering $25\%$ of the object width, or pasting a copy of the object shifted by $30\%$ of the image width). \Cref{tab:orberr_sanity} reports means over the 12 edited frames. OrbErr increases monotonically with the yaw and pitch offset on every object. Hue and brightness changes leave it near zero, below every geometric and structural edit. Missing and duplicated parts keep the camera exact, so RotErr is zero, whereas OrbErr assigns them errors comparable to a $5^\circ$ orbital offset; this is the correct-camera, wrong-object failure illustrated in \Cref{fig:rot_error_view_change}. RotErr also assigns identical errors to equal roll and yaw offsets, while the Euler term is exactly zero under roll and the full OrbErr is lower for roll than for yaw at $15^\circ$ and $30^\circ$. Finally, replacing the exact cameras with VGGT pose estimates, a different estimator from the Depth Anything~3 model used for the benchmark, yields $0.011$ on the unedited sequence, an estimator floor well below the differences in \Cref{tab:orberr_sanity}.

\begin{table}[!htbp]
\centering
\small
\caption{OrbErr sanity check on procedurally rendered objects with exact cameras. Means over three objects and four edited frames. RotErr is the per-frame geodesic error in degrees; the last column uses VGGT-estimated poses instead of exact cameras (unedited sequence: $0.011$).}
\label{tab:orberr_sanity}
\begin{tabular}{llcccc}
\toprule
Edit & Amount & RotErr ($^\circ$) & $E_{\mathrm{euler}}$ & OrbErr & OrbErr (VGGT) \\
\midrule
\multirow{3}{*}{Yaw} & $5^\circ$ & 5 & 0.014 & 0.114 & 0.124 \\
 & $15^\circ$ & 15 & 0.042 & 0.165 & 0.170 \\
 & $30^\circ$ & 30 & 0.083 & 0.219 & 0.229 \\
\midrule
\multirow{3}{*}{Pitch} & $5^\circ$ & 5 & 0.014 & 0.122 & 0.130 \\
 & $15^\circ$ & 15 & 0.042 & 0.164 & 0.169 \\
 & $30^\circ$ & 30 & 0.083 & 0.223 & 0.225 \\
\midrule
\multirow{3}{*}{Roll} & $5^\circ$ & 5 & 0.000 & 0.121 & 0.132 \\
 & $15^\circ$ & 15 & 0.000 & 0.150 & 0.159 \\
 & $30^\circ$ & 30 & 0.000 & 0.166 & 0.179 \\
\midrule
Hue & $+10^\circ$ & 0 & 0.000 & 0.003 & 0.015 \\
Brightness & $+10\%$ & 0 & 0.000 & 0.002 & 0.013 \\
Blur & $\sigma{=}2$\,px & 0 & 0.000 & 0.066 & 0.081 \\
\midrule
Missing part & $25\%$ & 0 & 0.000 & 0.108 & 0.122 \\
Duplicated object & $30\%$ shift & 0 & 0.000 & 0.122 & 0.139 \\
\bottomrule
\end{tabular}
\end{table}

\subsection{Pixel-Metric Sensitivity to Image Alignment}
\label{sec:supp_psnr_scale}
\begin{figure*}[!htbp]
    \centering
    \includegraphics[width=\linewidth,trim={0 0 0 22bp},clip]{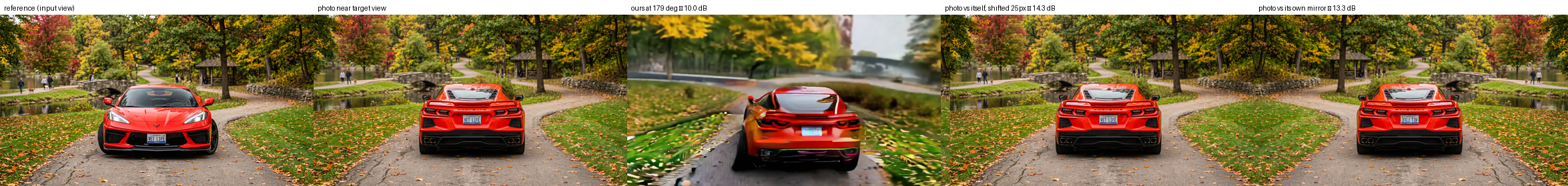}
    \caption{Pixel-metric sensitivity to alignment and orientation. Panels 1--3:
    reference, real near-endpoint photograph, and our $180^\circ$ output.
    Panels 4--5: the photograph translated by $3\%$ of its width
    ($14.3$\,dB) and horizontally reflected ($13.3$\,dB), scored against itself.
    The output-to-photograph score is illustrative: the real capture is not a
    pixel-aligned ground-truth endpoint.}
    \label{fig:psnr_scale}
\end{figure*}

Image fidelity and orbital fidelity capture complementary aspects of NVS.
\Cref{fig:psnr_scale} illustrates the sensitivity of pixel-aligned scores using
two transformations of the same photograph: a translation of $3\%$ of the
image width yields $14.3$\,dB PSNR, while horizontal reflection yields
$13.3$\,dB. These controls involve no synthesis and show how alignment and
orientation both affect the score. We therefore report target-view PSNR,
SSIM and LPIPS together with orbital and video-quality metrics.

\FloatBarrier
\section{Conditioning Sparsity and Orbital Range}
\label{sec:supp_sparsity}

These analyses test the mechanism motivating \ours{}: whether object
completion restores projected foreground support, and whether that gain
increases with orbital range. They measure conditioning coverage rather
than generated-image quality, complementing the synthesis results.

\subsection{Foreground Conditioning Support}
\label{sec:supp_sparsity_curve}
\label{sec:sparsity_analysis}
\begin{figure}[!htbp]
    \centering
    \includegraphics[width=\linewidth]{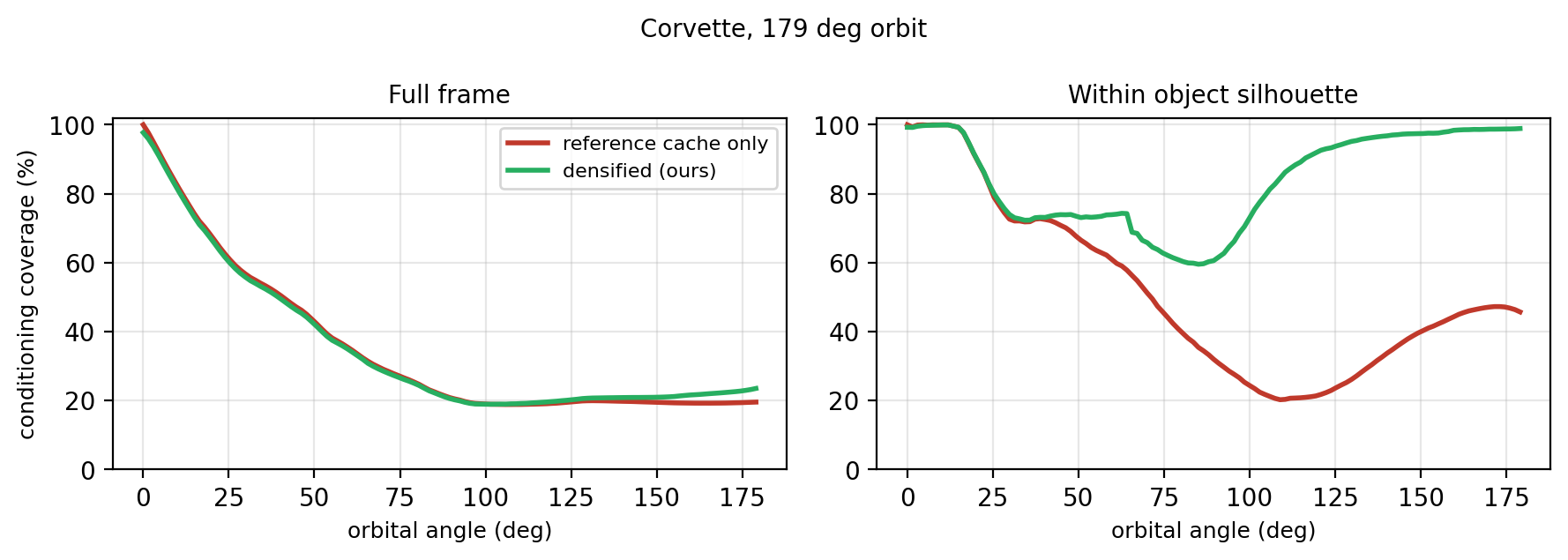}
    \caption{Conditioning coverage along the orbit for the reference cache
    (red) and densified cache (green). \textbf{Left:} full-frame coverage.
    \textbf{Right:} coverage within the rendered object footprint, increasing
    from $20.3\%$ to $84.5\%$ at the sparsest reference view near $109^\circ$.
    Densification supplies foreground support where the scene warp is sparse.}
    \label{fig:warp_sparsity}
\end{figure}

\Cref{fig:warp_sparsity} measures conditioning support along a $180^\circ$
orbit in one scene. Coverage is the fraction of target pixels receiving a
projected point, measured over the full frame and within the object-only
rendered footprint. We compare the reference cache with the densified cache
used by \ours{}.

At the sparsest foreground view, near $109^\circ$, densification increases
object coverage from $20.3\%$ to $84.5\%$, a factor of $4.2$.
At the endpoint, coverage rises from $45.7\%$ to $98.9\%$.
The full-frame minimum remains approximately $18.9\%$ near $106^\circ$,
showing that the added support is concentrated on the foreground object.
The largest gap occurs before the endpoint, as the originally visible surfaces
rotate out of view. These measurements quantify how object-centric completion
strengthens the foreground conditioning available to the generator.

\subsection{Effect of Orbital Range}
\label{sec:small_motion}
\begin{figure}[!htbp]
    \centering
    \includegraphics[width=\linewidth]{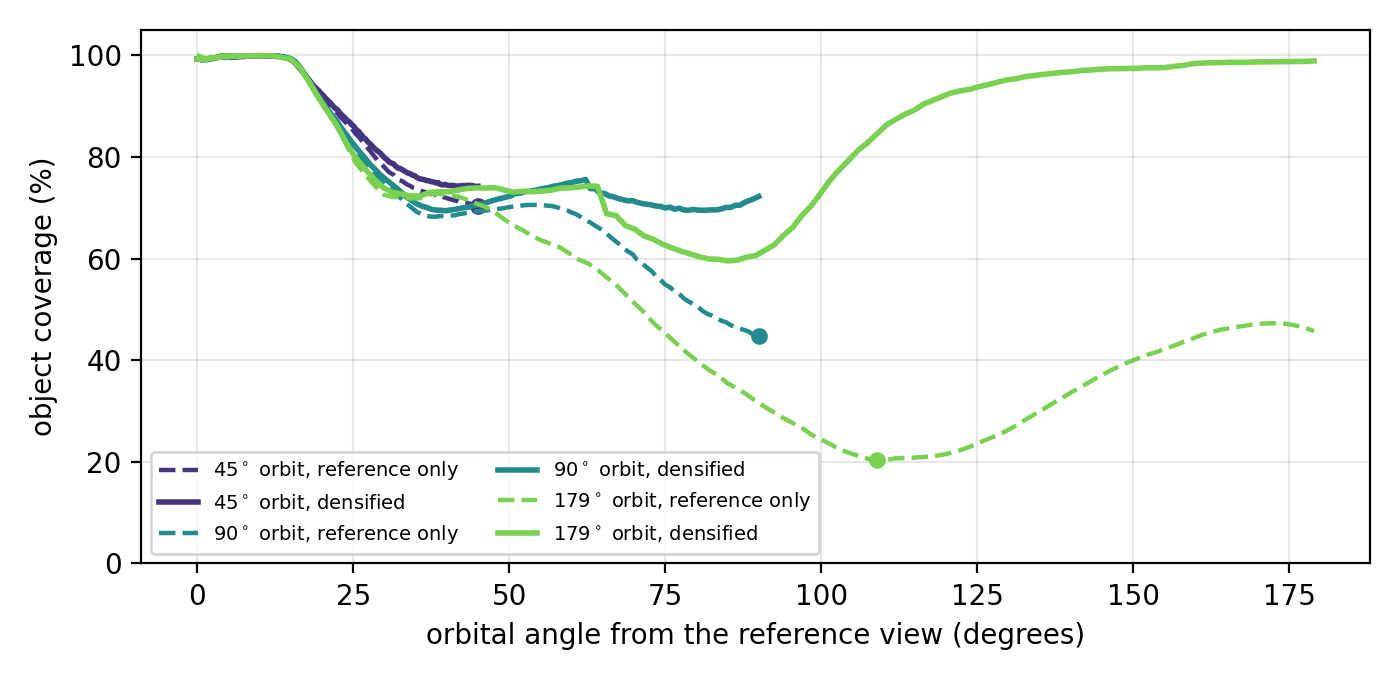}
    \caption{Effect of orbital range on foreground support. Dashed curves:
    reference-only cache; solid curves: densified cache. Markers identify the
    sparsest reference-only frame of each orbit. The coverage gain grows with
    orbital range. The later rise in reference-only coverage reflects
    reprojection of existing points, not newly reconstructed surfaces.}
    \label{fig:small_motion_coverage}
\end{figure}

We render the same scene over $45^\circ$, $90^\circ$ and $180^\circ$ orbits
to measure how the benefit of densification changes with viewpoint range
(\Cref{tab:small_motion_coverage,fig:small_motion_coverage}).
At each orbit's sparsest reference-only frame, object coverage increases from
$70.4\%$ to $74.2\%$, from $44.7\%$ to $72.2\%$, and from $20.3\%$ to
$84.5\%$, respectively. The widening coverage gap shows that densification
provides its largest support gain in the extreme-view regime targeted by
\ours{}.

\begin{table}[t]
    \centering
    \small
    \begin{tabular}{lccc}
        \toprule
        Orbital range & Worst angle & Reference-only & Densified \\
        \midrule
        $45^\circ$ & $45^\circ$ & $70.4\%$ & $74.2\%$ ($\times1.05$) \\
        $90^\circ$ & $90^\circ$ & $44.7\%$ & $72.2\%$ ($\times1.62$) \\
        $179^\circ$ & $109^\circ$ & $20.3\%$ & $84.5\%$ ($\times4.17$) \\
        \bottomrule
    \end{tabular}
    \caption{Conditioning coverage within the rendered object footprint at
    the sparsest reference-only frame of each orbit, for the scene of
    \Cref{fig:warp_sparsity}. The relative support gain from densification
    increases from $1.05\times$ at $45^\circ$ to $4.17\times$ at $179^\circ$.}
    \label{tab:small_motion_coverage}
\end{table}

\FloatBarrier
\section{Additional Quantitative Results}
\label{sec:supp_quant}

We provide a per-dataset breakdown of the single-view novel view synthesis benchmark in \Cref{tab:nvs_results_sv}, complementing the aggregated results reported in the main paper. The table reports image similarity (PSNR, SSIM, LPIPS), semantic consistency (CLIP-F), camera pose accuracy (RotErr, TransErr), feature matching (MatchPx), and VBench video-quality metrics (TempFlick, SubjCons, BGCons, ImagQual, AesQual, MotSmooth) across the five evaluation datasets: CO3D, Objectron, 360\textdegree{}, OmniObject3D, and SynView-X.

Across image-similarity metrics, \ours{} achieves the best LPIPS on all five
datasets and the best PSNR on four of five. On SynView-X, \ours{} also
achieves the best SSIM.

\begin{table*}[t]
\centering
\caption{
  Novel view synthesis benchmark results (single-view setting, no multi-view conditioning).
  $\uparrow$ higher is better, $\downarrow$ lower is better.
}
\label{tab:nvs_results_sv}
\resizebox{0.93\textwidth}{!}{%
\begin{tabular}{l l ccccccc}
\toprule
Dataset & Metric
  & TrajCr. & GEN3C & ReCamM. & EX-4D & CogNVS & SEVA & \ours \\
\midrule

\multirow{13}{*}{\rotatebox{90}{CO3D}}
& PSNR$\uparrow$       & 5.86  & 6.52  & 3.92  & 5.25  & 5.70  & 6.42  & \B{10.45} \\
& SSIM$\uparrow$       & 0.416 & 0.394 & 0.407 & 0.412 & 0.472 & 0.439 & \B{0.581} \\
& LPIPS$\downarrow$    & 0.819 & 0.798 & 0.942 & 0.834 & 0.810 & 0.798 & \B{0.429} \\
& CLIP-F$\uparrow$     & 93.53 & 93.80 & 93.62 & 95.52 & 94.10 & \B{95.90} & 94.97 \\
& RotErr$\downarrow$   & 173.0 & 172.5 & 156.0 & 178.1 & 160.9 & 155.8 & \B{140.27} \\
& TransErr$\downarrow$ & 0.035 & 0.029 & 0.049 & 0.022 & 0.023 & \B{0.016} & 0.022 \\
& MatchPx$\uparrow$    & 0.004 & \B{0.009} & 0.001 & 0.006 & 0.004 & 0.008 & 0.008 \\
& TempFlick$\uparrow$  & 0.931 & 0.948 & \B{0.971} & 0.952 & 0.959 & 0.958 & 0.955 \\
& SubjCons$\uparrow$   & 0.806 & 0.753 & 0.710 & 0.789 & 0.807 & \B{0.869} & 0.829 \\
& BGCons$\uparrow$     & 0.908 & 0.889 & 0.848 & 0.887 & 0.871 & \B{0.934} & 0.909 \\
& ImagQual$\uparrow$   & 0.565 & 0.519 & 0.357 & \B{0.579} & 0.565 & 0.560 & 0.496 \\
& AesQual$\uparrow$    & 0.953 & 0.784 & 0.453 & 0.978 & 0.884 & \B{0.986} & 0.445 \\
& MotSmooth$\uparrow$  & 0.384 & 0.691 & \B{0.916} & 0.619 & 0.624 & 0.659 & 0.775 \\
\midrule

\multirow{13}{*}{\rotatebox{90}{Objectron}}
& PSNR$\uparrow$       & 10.43 & 9.19  & 14.59 & 11.50 & 11.62 & 9.51  & \B{15.14} \\
& SSIM$\uparrow$       & 0.675 & 0.649 & \B{0.783} & 0.698 & 0.712 & 0.656 & 0.714 \\
& LPIPS$\downarrow$    & 0.765 & 0.741 & 0.774 & 0.750 & 0.738 & 0.731 & \B{0.377} \\
& CLIP-F$\uparrow$     & 94.14 & 94.19 & 93.62 & \B{95.53} & 94.92 & 95.06 & 94.36 \\
& RotErr$\downarrow$   & 313.4 & 285.8 & 261.5 & 300.0 & 300.3 & 311.0 & \B{11.76} \\
& TransErr$\downarrow$ & 0.076 & 0.087 & 0.103 & 0.046 & 0.058 & 0.028 & \B{0.018} \\
& MatchPx$\uparrow$    & 0.001 & \B{0.004} & 0.000 & 0.001 & 0.001 & 0.002 & 0.003 \\
& TempFlick$\uparrow$  & 0.949 & \B{0.968} & 0.928 & 0.964 & 0.965 & 0.962 & 0.966 \\
& SubjCons$\uparrow$   & 0.801 & 0.770 & 0.651 & 0.787 & 0.804 & \B{0.839} & 0.759 \\
& BGCons$\uparrow$     & \B{0.916} & 0.913 & 0.863 & 0.900 & 0.888 & 0.828 & 0.908 \\
& ImagQual$\uparrow$   & 0.502 & 0.462 & 0.276 & 0.467 & 0.471 & \B{0.523} & 0.498 \\
& AesQual$\uparrow$    & 0.456 & 0.228 & 0.111 & \B{0.484} & 0.300 & 0.469 & 0.343 \\
& MotSmooth$\uparrow$  & 0.532 & 0.854 & \B{0.976} & 0.752 & 0.797 & 0.663 & 0.784 \\
\midrule

\multirow{13}{*}{\rotatebox{90}{360\textdegree{}}}
& PSNR$\uparrow$       & 4.71  & 4.31  & 3.39  & 4.62  & 5.25  & 5.40  & \B{11.26} \\
& SSIM$\uparrow$       & 0.228 & 0.208 & 0.239 & 0.182 & 0.332 & 0.278 & \B{0.582} \\
& LPIPS$\downarrow$    & 0.894 & 0.878 & 0.964 & 0.927 & 0.846 & 0.850 & \B{0.421} \\
& CLIP-F$\uparrow$     & 92.66 & 94.31 & 93.95 & 94.13 & 93.32 & \B{94.92} & 93.74 \\
& RotErr$\downarrow$   & 178.9 & 249.2 & 169.9 & 250.4 & 221.8 & 158.9 & \B{133.31} \\
& TransErr$\downarrow$ & 0.038 & 0.061 & 0.071 & 0.041 & \B{0.030} & \B{0.030} & 0.050 \\
& MatchPx$\uparrow$    & 0.005 & 0.007 & 0.001 & 0.005 & 0.005 & \B{0.009} & 0.007 \\
& TempFlick$\uparrow$  & 0.896 & 0.923 & \B{0.956} & 0.916 & 0.930 & 0.912 & 0.924 \\
& SubjCons$\uparrow$   & 0.801 & 0.763 & 0.724 & 0.762 & 0.768 & \B{0.826} & 0.748 \\
& BGCons$\uparrow$     & 0.894 & 0.913 & 0.885 & 0.881 & 0.877 & \B{0.921} & 0.895 \\
& ImagQual$\uparrow$   & 0.504 & 0.476 & 0.364 & \B{0.654} & 0.501 & 0.584 & 0.490 \\
& AesQual$\uparrow$    & \B{0.999} & 0.841 & 0.741 & 0.908 & 0.758 & 0.996 & 0.403 \\
& MotSmooth$\uparrow$  & 0.144 & 0.445 & \B{0.641} & 0.162 & 0.264 & 0.312 & 0.496 \\
\midrule

\multirow{13}{*}{\rotatebox{90}{\shortstack{Omni\\Object3D}}}
& PSNR$\uparrow$       & 5.70  & 4.75  & 4.49  & 5.43  & 5.90  & 7.07   & \B{9.36} \\
& SSIM$\uparrow$       & 0.511 & 0.412 & 0.440 & 0.475 & 0.511 & \B{0.601}   & 0.482 \\
& LPIPS$\downarrow$    & 0.801 & 0.794 & 0.839 & 0.775 & 0.779 & 0.656   & \B{0.537} \\
& CLIP-F$\uparrow$     & 94.51 & 95.05 & 93.95 & 94.87 & \B{95.29} & 95.06   & 94.78 \\
& RotErr$\downarrow$   & 169.6 & 171.5 & 160.3 & 155.6 & 148.9 & 152.2   & \B{141.08} \\
& TransErr$\downarrow$ & 0.109 & 0.150 & 0.113 & 0.141 & 0.091 & 0.052   & \B{0.013} \\
& MatchPx$\uparrow$    & 0.002 & 0.004 & 0.000 & 0.002 & 0.002 & \B{0.008}   & 0.004 \\
& TempFlick$\uparrow$  & 0.953 & 0.972 & \B{0.981} & 0.967 & 0.975 & 0.962   & 0.971 \\
& SubjCons$\uparrow$   & 0.810 & 0.768 & 0.741 & 0.707 & 0.830 & \B{0.840}   & 0.793 \\
& BGCons$\uparrow$     & 0.908 & 0.908 & 0.882 & 0.877 & 0.908 & \B{0.919}   & 0.909 \\
& ImagQual$\uparrow$   & 0.536 & 0.489 & 0.369 & 0.403 & 0.522 & \B{0.624}   & 0.530 \\
& AesQual$\uparrow$    & \B{0.932} & 0.846 & 0.609 & 0.823 & 0.860 & 0.930   & 0.450 \\
& MotSmooth$\uparrow$  & 0.549 & 0.857 & \B{0.896} & 0.576 & 0.791 & 0.677   & 0.881 \\
\midrule

\multirow{13}{*}{\rotatebox{90}{SynView-X}}
& PSNR$\uparrow$       & 11.19 & 12.50   & \B{13.18} & 12.61 & 11.93 & 10.00   & 12.14 \\
& SSIM$\uparrow$       & 0.470 & 0.417   & 0.537 & 0.555 & 0.538 & 0.428   & \B{0.586} \\
& LPIPS$\downarrow$    & 0.676 & 0.624   & 0.734 & 0.666 & 0.665 & 0.644   & \B{0.456} \\
& CLIP-F$\uparrow$     & 93.25 & 94.63   & 91.95 & \B{95.85} & 93.66 & 94.57   & 93.85 \\
& RotErr$\downarrow$   & 54.37 & 41.51   & 69.73 & 66.64 & 50.66 & 58.50   & \B{29.63} \\
& TransErr$\downarrow$ & 0.086 & 0.137   & 0.173 & 0.165 & 0.113 & 0.107   & \B{0.017} \\
& MatchPx$\uparrow$    & 0.003 & 0.007   & 0.000 & 0.002 & 0.003 & \B{0.009}   & 0.008 \\
& TempFlick$\uparrow$  & 0.951 & 0.972   & \B{0.979} & 0.974 & 0.974 & 0.953   & 0.964 \\
& SubjCons$\uparrow$   & 0.773 & 0.797   & 0.648 & 0.736 & 0.788 & \B{0.803}   & 0.781 \\
& BGCons$\uparrow$     & 0.901 & \B{0.920}   & 0.875 & 0.901 & 0.884 & 0.915   & 0.899 \\
& ImagQual$\uparrow$   & 0.421 & 0.440   & 0.245 & 0.346 & 0.385 & \B{0.547}   & 0.421 \\
& AesQual$\uparrow$    & 0.723 & 0.735   & 0.083 & 0.773 & 0.635 & \B{0.883}   & 0.392 \\
& MotSmooth$\uparrow$  & 0.463 & 0.849   & \B{0.883} & 0.721 & 0.761 & 0.610   & 0.789 \\

\bottomrule
\end{tabular}
}
\end{table*}

\paragraph{Video quality.}
\Cref{tab:nvs_mean_sv_vbench,tab:nvs_mean_mv_vbench} report full-frame
VBench metrics~\citep{huang2024vbench} for the single-view and auxiliary-view
settings, averaged across the five datasets. These assess temporal, subject and
background consistency, per-frame image quality, and motion smoothness,
complementing the object-region and camera-motion metrics.

\begin{table*}[!htbp]
\centering
\caption{{\bf Single-view} VBench video-quality metrics averaged across five datasets. OrbErr is reported on SynView-X only, matching \Cref{tab:nvs_mean_sv_main}; \ours{}-GEN3C has the lowest reported OrbErr.}
\label{tab:nvs_mean_sv_vbench}
\footnotesize
\setlength{\tabcolsep}{6pt}
\renewcommand{\arraystretch}{1.25}
\adjustbox{max width=\linewidth}{%
\begin{tabular}{c l ccccc c}
\toprule
& & \multicolumn{3}{c}{Consistency} & \multicolumn{1}{c}{Quality} & \multicolumn{1}{c}{Motion} & \multicolumn{1}{c}{Camera Pose} \\
\cmidrule(lr){3-5} \cmidrule(lr){6-6} \cmidrule(lr){7-7} \cmidrule(lr){8-8}
& Method
    & TempFlick$\uparrow$ & SubjCons$\uparrow$ & BGCons$\uparrow$
    & ImagQual$\uparrow$
    & MotSmooth$\uparrow$
    & OrbErr$\downarrow$ \\
\midrule
\multirow{5}{*}{\scriptsize\shortstack{Warp-\\based}}
    & TrajCr.       & 0.936 & 0.798 & 0.905 & 0.506 & 0.414 & 0.3022 \\
    & GEN3C         & 0.957 & 0.770 & \B{0.909} & 0.477 & 0.739 & 0.3058 \\
    & EX-4D         & 0.955 & 0.756 & 0.889 & 0.490 & 0.566 & 0.3192 \\
    & CogNVS        & 0.961 & 0.799 & 0.886 & 0.489 & 0.647 & 0.3063 \\
    & WF            & \B{0.965} & 0.779 & 0.888 & 0.484 & 0.577 & 0.3067 \\
\cmidrule(lr){2-8}
\multirow{2}{*}{\scriptsize\shortstack{Camera-\\cond.}}
    & ReCamM.       & 0.963 & 0.695 & 0.871 & 0.322 & \B{0.862} & 0.3460 \\
    & SEVA          & 0.949 & \B{0.835} & 0.903 & 0.568 & 0.584 & 0.2855 \\
\cmidrule(lr){2-8}
    & \ours-TC      & 0.931 & 0.786 & 0.891 & \B{0.569} & 0.464 & 0.2288 \\
    & \ours-GEN3C   & 0.956 & 0.782 & 0.904 & 0.487 & 0.745 & \B{0.2045} \\
\bottomrule
\end{tabular}}
\end{table*}

\begin{table*}[!htbp]
\centering
\caption{{\bf Auxiliary-view} VBench video-quality metrics across 5 datasets (CO3D, Objectron, 360\textdegree{}, OmniObject3D, SynView-X). \textcolor{gray}{Gray} methods use GT camera poses and in-scene second views and are shown for reference only.}
\label{tab:nvs_mean_mv_vbench}
\footnotesize
\setlength{\tabcolsep}{6pt}
\renewcommand{\arraystretch}{1.25}
\begin{tabular}{c l ccc c c}
\toprule
& & \multicolumn{3}{c}{Consistency} & \multicolumn{1}{c}{Quality} & \multicolumn{1}{c}{Motion} \\
\cmidrule(lr){3-5} \cmidrule(lr){6-6} \cmidrule(lr){7-7}
& Method & TempFlick$\uparrow$ & SubjCons$\uparrow$ & BGCons$\uparrow$ & ImagQual$\uparrow$ & MotSmooth$\uparrow$ \\
\midrule
\multirow{2}{*}{\scriptsize\shortstack{w/ GT}}
& \textcolor{gray}{GEN3C-MV} & \textcolor{gray}{0.951} & \textcolor{gray}{0.755} & \textcolor{gray}{0.894} & \textcolor{gray}{0.433} & \textcolor{gray}{0.642} \\
& \textcolor{gray}{SEVA-MV}  & \textcolor{gray}{0.954} & \textcolor{gray}{0.812} & \textcolor{gray}{0.904} & \textcolor{gray}{0.576} & \textcolor{gray}{0.501} \\
\cmidrule(lr){2-7}
\multirow{3}{*}{\scriptsize\shortstack{w/o GT}}
& GEN3C-MV & 0.955 & 0.769 & 0.899 & 0.426 & 0.684 \\
& SEVA-MV  & 0.938 & 0.769 & 0.886 & \B{0.581} & 0.442 \\
& \ours-GEN3C              & \B{0.966} & \B{0.785} & \B{0.905} & 0.471 & \B{0.835} \\
\bottomrule
\end{tabular}
\end{table*}

\FloatBarrier
\section{Single-Image Comparison with FreeOrbit4D}
\label{sec:supp_freeorbit4d}

FreeOrbit4D~\citep{freeorbit4d} and \ours{} both render completed geometry to
guide video diffusion, but obtain that geometry differently. FreeOrbit4D
generates multiview frames with SV4D from an input video and fuses them into a
4D proxy; \ours{} uses a native 3D generative prior applied to a single image.
We compare the methods on a static reference image to examine geometry
completion without inter-frame parallax.

\paragraph{Protocol.}
We repeat the reference image into a 45-frame $854\times480$ clip and supply
the same foreground mask used by \ours{}, retaining FreeOrbit4D's released
configuration, checkpoints and seed. Both trajectories use a $180^\circ$ yaw.

\begin{figure*}[!htbp]
    \centering
    {\small Input video}\\
    \includegraphics[width=\linewidth]{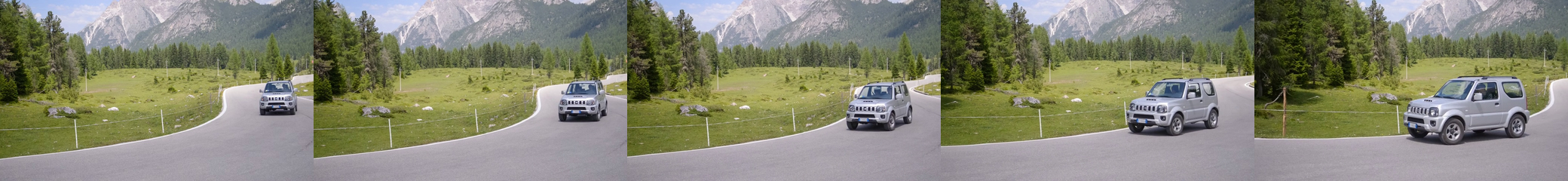}\\
    {\small 4D proxy render}\\
    \includegraphics[width=\linewidth]{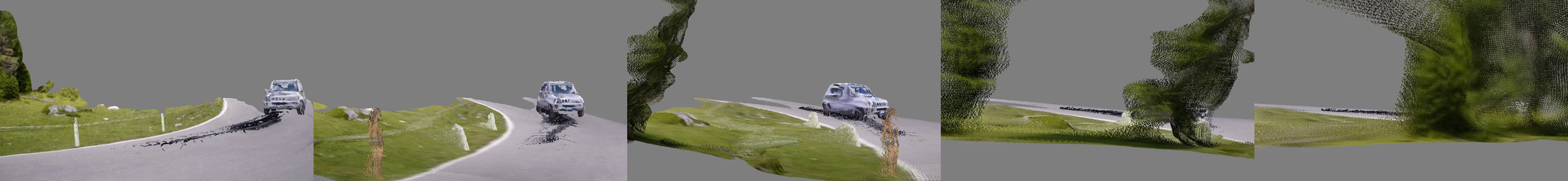}\\
    {\small FreeOrbit4D output}\\
    \includegraphics[width=\linewidth]{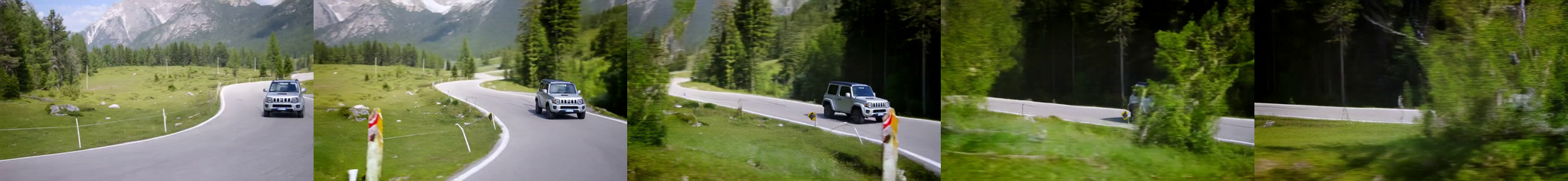}
    \caption{FreeOrbit4D on the authors' bundled \texttt{car-turn} video using
    the released configuration. Rows: input video, 4D proxy render and generated
    output. Motion-rich input supports a dense proxy; foreground vegetation
    occludes the vehicle in some views.}
    \label{fig:freeorbit4d_install_check}
\end{figure*}

\paragraph{Native-video reference.}
We also run the authors' bundled \texttt{car-turn} video with their released
configuration (\Cref{fig:freeorbit4d_install_check}). Its proxy remains dense,
with projected coverage changing from $49.0\%$ to $39.1\%$ along the orbit.
The generated sequence follows the proxy, including vegetation occlusions.

\begin{figure*}[!htbp]
    \centering
    \ifdefined\foroww\else\newlength{\foroww}\newlength{\focellw}\fi
    \setlength{\foroww}{0.94\linewidth}
    \setlength{\focellw}{0.188\linewidth}
    \setlength{\tabcolsep}{1pt}
    \renewcommand{\arraystretch}{1.15}
    \begin{tabular}{@{}c@{\hspace{3pt}}c@{}}
        & \makebox[\foroww]{%
            \makebox[\focellw]{\small $0^\circ$}%
            \makebox[\focellw]{\small $45^\circ$}%
            \makebox[\focellw]{\small $90^\circ$}%
            \makebox[\focellw]{\small $135^\circ$}%
            \makebox[\focellw]{\small $180^\circ$}} \\
        \rotatebox[origin=c]{90}{\shortstack{\small \ours{}\\\small output}} &
            \raisebox{-0.5\height}{\includegraphics[width=\foroww]{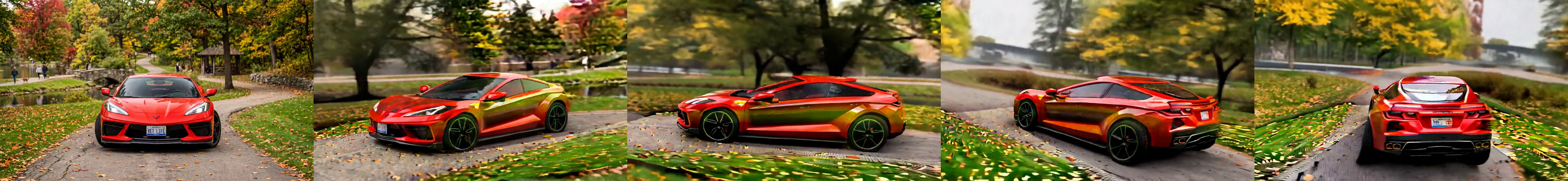}} \\
        \rotatebox[origin=c]{90}{\shortstack{\small FreeOrbit4D\\\small proxy}} &
            \raisebox{-0.5\height}{\includegraphics[width=\foroww]{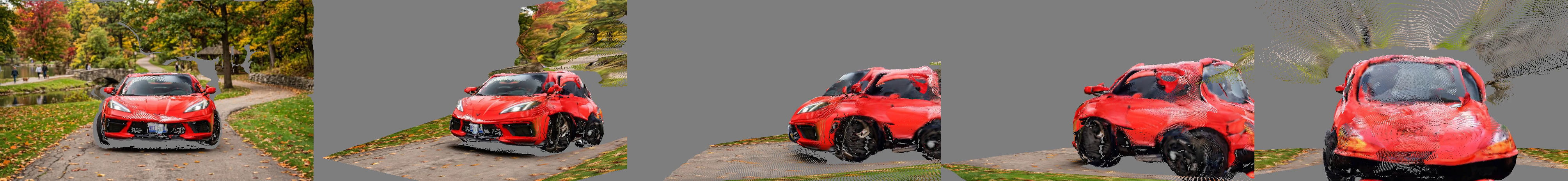}} \\
        \rotatebox[origin=c]{90}{\shortstack{\small FreeOrbit4D\\\small output}} &
            \raisebox{-0.5\height}{\includegraphics[width=\foroww]{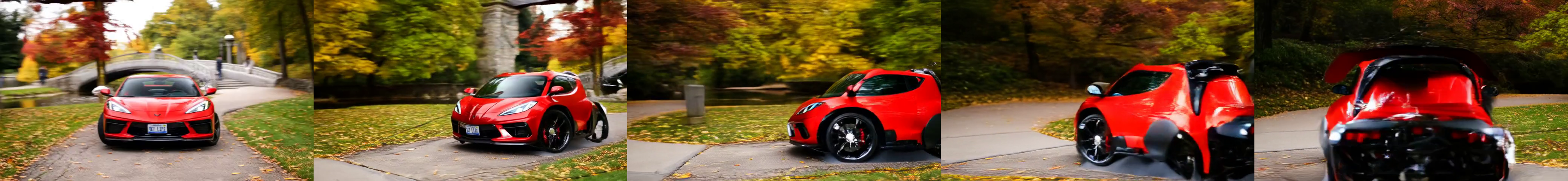}} \\
    \end{tabular}
    \caption{Single-image comparison at corresponding azimuths.
    Rows: \ours{} output, FreeOrbit4D proxy and FreeOrbit4D output.
    FreeOrbit4D receives a repeated reference image; \ours{} uses the image
    directly. In this example, object-prior completion preserves the car's
    proportions as the opposite side comes into view.}
    \label{fig:freeorbit4d_comparison}
\end{figure*}

\paragraph{Single-image results.}
On the static-input example in \Cref{fig:freeorbit4d_comparison}, FreeOrbit4D's
proxy coverage falls from $96.6\%$ to $32.6\%$. The far-side reconstruction
becomes sparse, and the generated car develops a shorter wheelbase and an
altered roofline. \ours{} retains the car's silhouette and proportions across
the orbit. This example illustrates the benefit of a native 3D object prior:
it supplies unobserved surface geometry without relying on parallax between
input frames.

\FloatBarrier
\section{Additional Qualitative Results}
\label{sec:supp_qual}

For additional visual examples, including cars, furniture, auxiliary-view fusion
and multi-object scenes, please see the project page.

\begin{figure*}[!htbp]
    \centering
    \includegraphics[width=\linewidth]{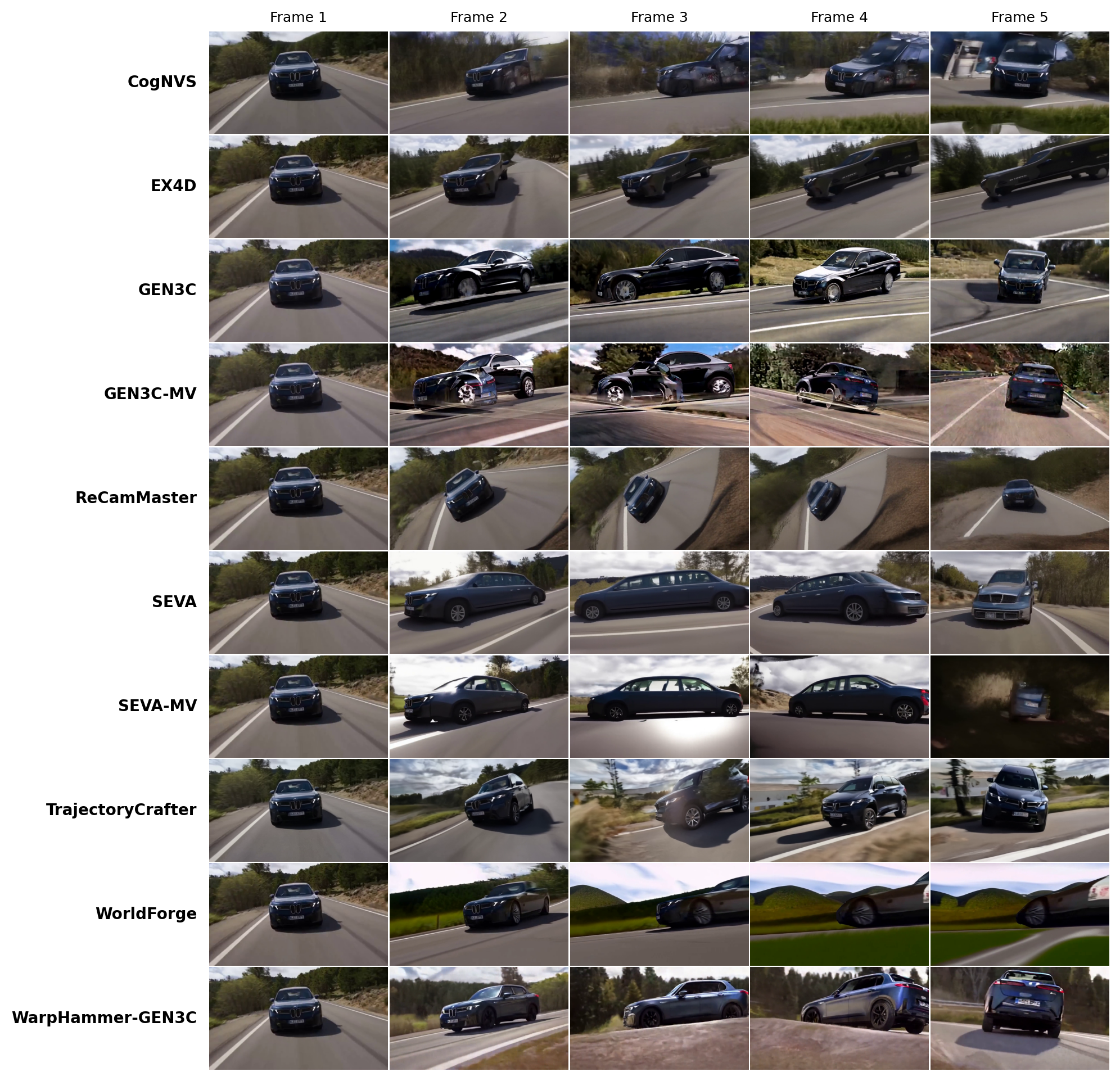}
    \caption{Novel view synthesis comparison along a full orbital trajectory around a car. Warp-based baselines exhibit ghosting and mirror-like duplication at large angles, camera-conditioned baselines fail to follow the orbital motion, while \ours{} produces a coherent, geometry-faithful orbital sequence.}
    \label{fig:qual_supp_car}
\end{figure*}

We present additional qualitative comparisons that complement the main paper. \Cref{fig:qual_supp_car} shows novel view synthesis results with auxiliary view, where applicable, from the back of the car showing a full orbital trajectory around a car and comparing all baselines side-by-side with \ours{}. The example is selected to span a wide range of azimuthal viewpoints, including angles far from the input view, where the differences between methods become most apparent.

Warp-based baselines (TrajectoryCrafter, GEN3C, EX-4D, CogNVS) reproduce the input view faithfully at small offsets but progressively degrade as the orbital angle grows: occluded regions of the car are filled with warped ghost geometry, and large-angle frames frequently exhibit mirror-like duplications of the visible side rather than the genuine far side of the object. Camera-conditioned baselines (ReCamMaster, SEVA) avoid such warping artifacts but instead produce near-static syntheses that fail to follow the prescribed orbital trajectory, so the object viewpoint barely changes across frames. In contrast, \ours{} produces a coherent orbital sequence in which the car is rendered from the correct viewpoint at each frame, with consistent geometry, plausible far-side appearance, and stable background, demonstrating the benefit of object-centric 3D densification under large viewpoint changes.

\FloatBarrier

\begin{figure*}[!htbp]
\centering
{\small Sofa}\\
\begin{tabular}{cc}
Reference & Auxiliary\\
\includegraphics[width=0.30\linewidth]{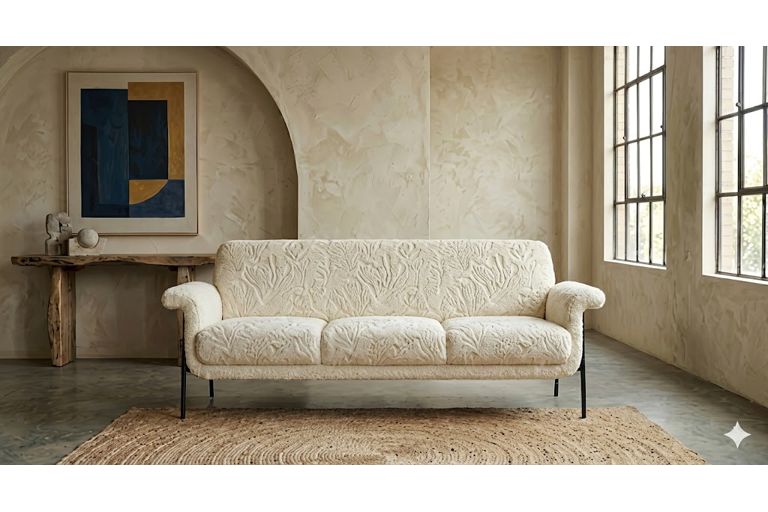} &
\includegraphics[width=0.30\linewidth]{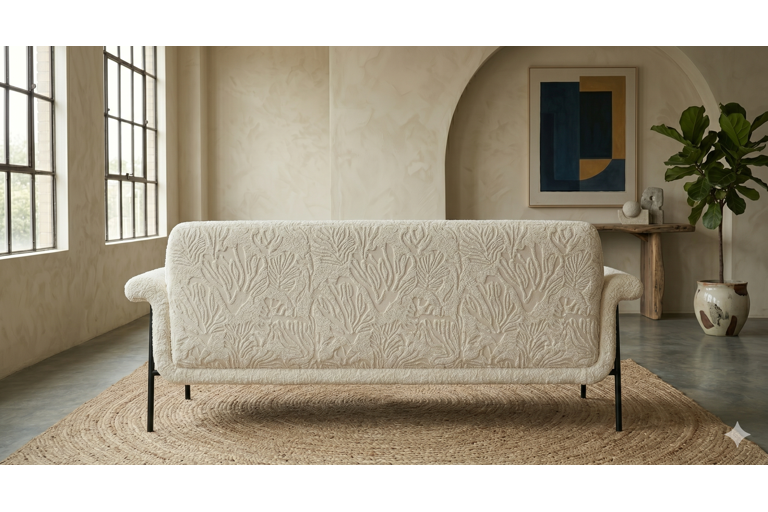}
\end{tabular}\\
\includegraphics[width=\linewidth]{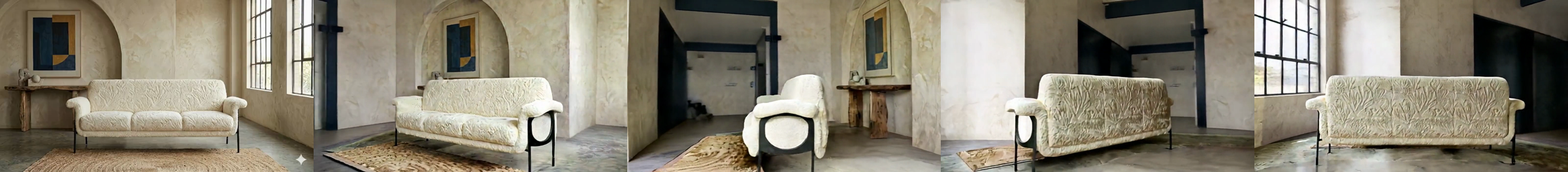}\\[1ex]
{\small Office chair}\\
\begin{tabular}{cc}
Reference & Auxiliary\\
\includegraphics[width=0.30\linewidth]{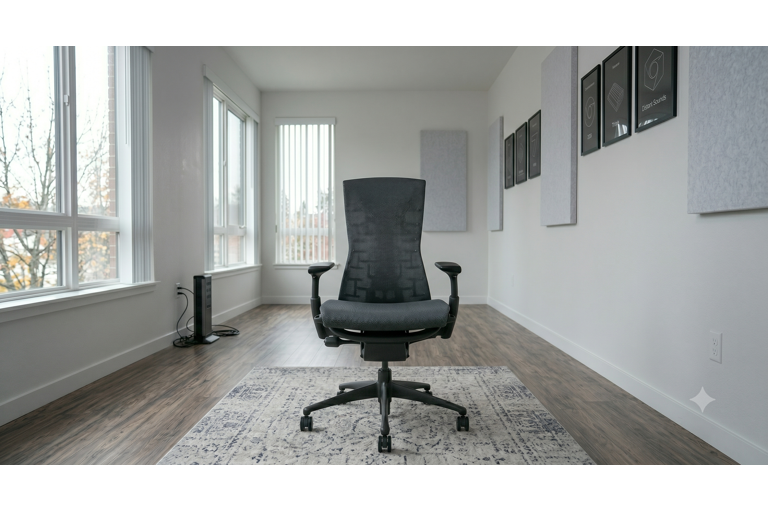} &
\includegraphics[width=0.30\linewidth]{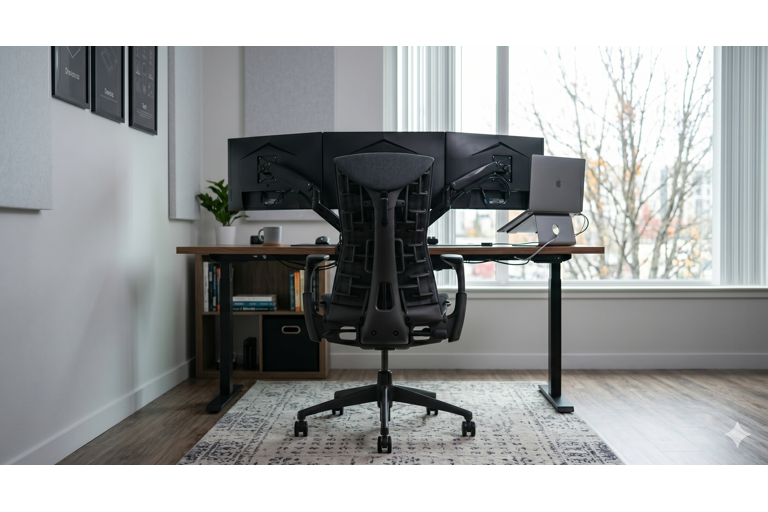}
\end{tabular}\\
\includegraphics[width=\linewidth]{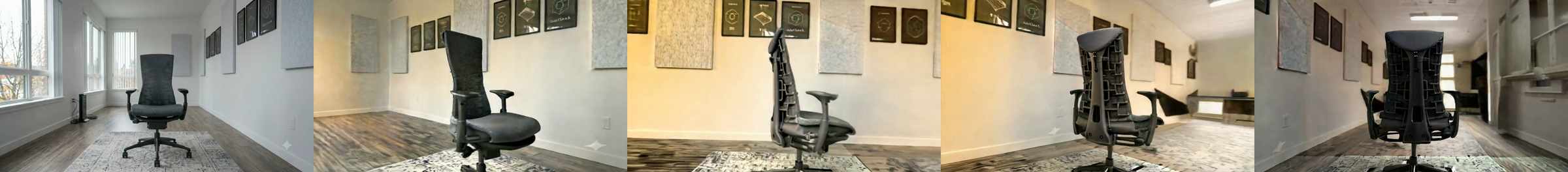}
\caption{Additional auxiliary-view synthesis examples. Each output row samples
the corresponding project-page video at five uniformly spaced times, from the
initial view to the final view. Inputs and output frames are shown without
cropping.}
\label{fig:qual_website_objects}
\end{figure*}

\begin{figure*}[!htbp]
\centering
{\small Reference scene 1}\\
\begin{tabular}{cc}
Reference & Auxiliary\\
\includegraphics[width=0.30\linewidth]{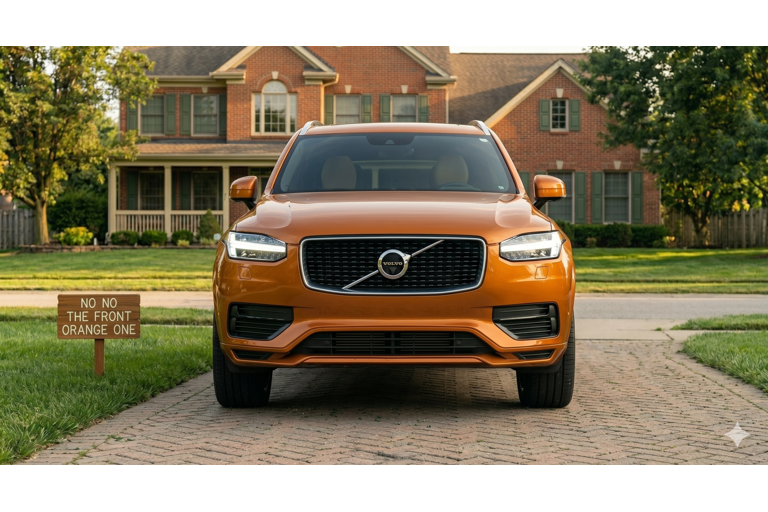} &
\includegraphics[width=0.30\linewidth]{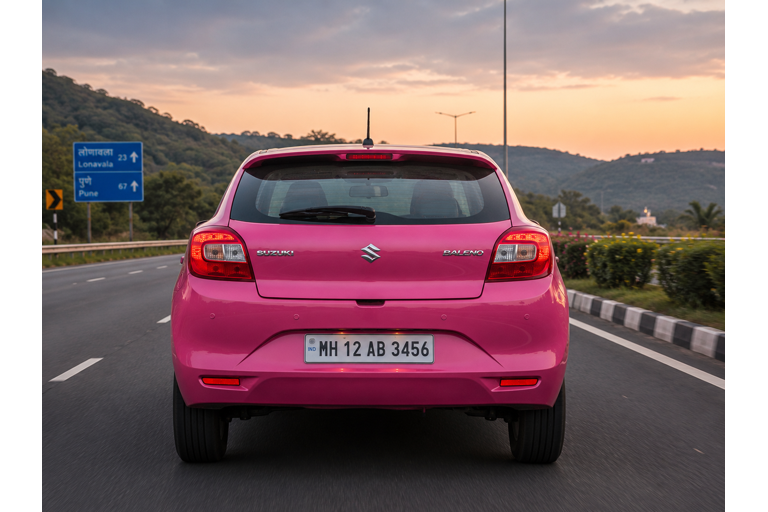}
\end{tabular}\\
\includegraphics[width=\linewidth]{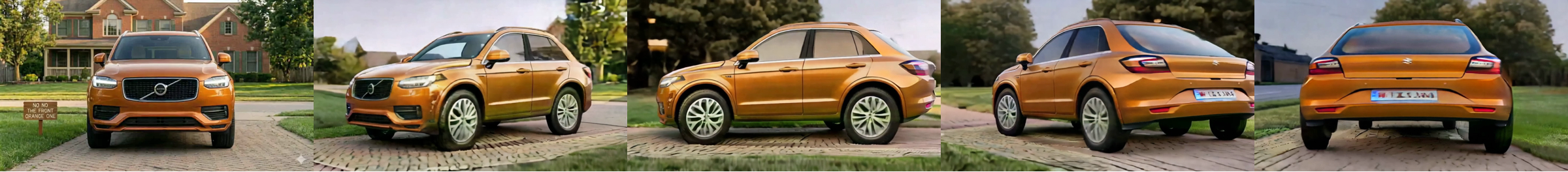}\\[1ex]
{\small Reference scene 2}\\
\begin{tabular}{cc}
Reference & Auxiliary\\
\includegraphics[width=0.30\linewidth]{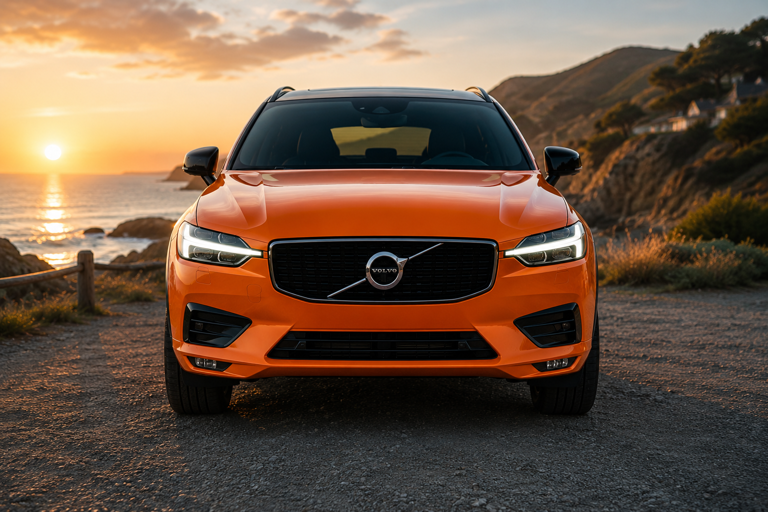} &
\includegraphics[width=0.30\linewidth]{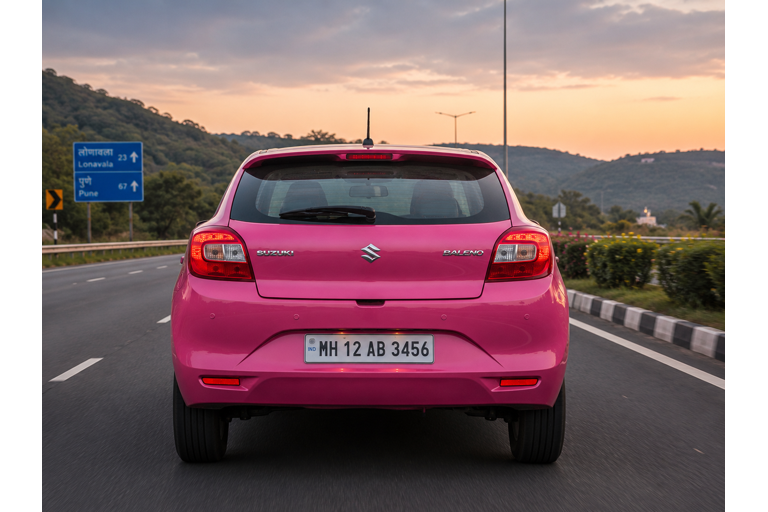}
\end{tabular}\\
\includegraphics[width=\linewidth]{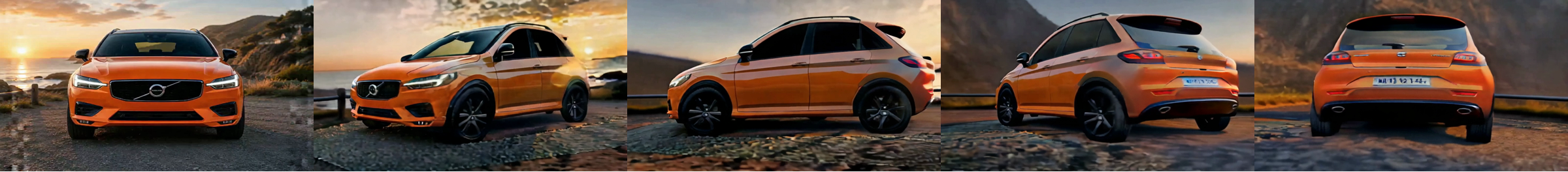}
\caption{Cross-instance auxiliary fusion in two reference scenes. An orange
reference car is paired with the rear view of a different, pink car. The output
rows show five uniformly spaced frames from the project-page fusion videos,
illustrating reference appearance retained during the synthesized orbit.}
\label{fig:qual_website_fusion}
\end{figure*}

\paragraph{Further auxiliary-view examples.}
\Cref{fig:qual_website_objects} adds sofa and office-chair sequences from the
project-page gallery, showing the reference, auxiliary and five uniformly
spaced output frames. \Cref{fig:qual_website_fusion} shows two cross-instance
car-fusion examples using the same auxiliary rear view with different reference
scenes. The generated cars retain the orange reference appearance despite the
pink auxiliary. Full videos accompany these qualitative examples.

\end{document}